\documentclass[11pt]{article}

\newif\ifarxiv
\arxivtrue
\newif\ifshowrev
\newcommand{\rev}[1]{%
  \ifshowrev
    \textcolor{blue}{#1}%
  \else
    #1%
  \fi
}

\usepackage[final]{acl}

\usepackage{times}
\usepackage{latexsym}
\usepackage{booktabs}
\usepackage{hyperref}

\usepackage[T1]{fontenc}
\usepackage[utf8]{inputenc}

\usepackage{microtype}

\usepackage{inconsolata}

\usepackage{graphicx}
\usepackage[hyperfirst=false, nonumberlist, nostyles, nogroupskip]{glossaries}
\glsdisablehyper
\usepackage{fontawesome5}
\usepackage{pgfplots}
\pgfplotsset{compat=1.18}

\usepackage{xcolor}

\ifarxiv
  \usepackage{listings}
  \usepackage{upquote}
  \usepackage{xkeyval}

\lstdefinestyle{gradiendcode}{
  language=Python,
  basicstyle=\ttfamily\small,
  breaklines=true,
  frame=lines,
  framesep=2mm,
  columns=fullflexible,
  keepspaces=true,
  showstringspaces=false,
  upquote=true,
  tabsize=2,
  keywordstyle=\color{blue}\bfseries,
  commentstyle=\color{gray}\itshape,
  stringstyle=\color{teal},
  literate={id}{{id}}2,
aboveskip=0.6\baselineskip,
  belowskip=0.6\baselineskip,
  prebreak=\raisebox{0ex}[0ex][0ex]{},
  postbreak=\raisebox{0ex}[0ex][0ex]{},
}
  \makeatletter
  \define@key{minted}{fontsize}{\def\minted@fontsize{#1}}
  \define@key{minted}{breaklines}[true]{\lstset{breaklines=true}}
  \define@key{minted}{frame}{\lstset{frame=#1}}
  \define@key{minted}{framesep}{\lstset{framesep=#1}}
  \define@key{minted}{autogobble}[true]{}
  \define@key{minted}{linenos}[true]{\lstset{numbers=left,numberstyle=\tiny,numbersep=6pt}}

  \def\minted@fontsize{\small}

  \newcommand{\setminted}[1]{%
    \def\minted@fontsize{\small}%
    \setkeys{minted}{#1}%
    \lstset{style=gradiendcode,basicstyle=\ttfamily\minted@fontsize}%
  }

  \lstnewenvironment{minted}[2][]{%
    \def\minted@fontsize{\small}%
    \setkeys{minted}{#1}%
    \lstset{style=gradiendcode,language=#2,basicstyle=\ttfamily\minted@fontsize}%
  }{}
  \makeatother

\else
  \usepackage{minted}
  \setminted{
    fontsize=\small,
    breaklines=true,
    autogobble=true,
    frame=lines,
    framesep=2mm
  }
\fi

\newcommand{\gradiend}{\textsc{Gradiend}}

\newacronym{gradiend}{\gradiend}{GRADIent ENcoder Decoder}

\newacronym{mlm}{MLM}{Masked Language Modeling}
\newacronym{clm}{CLM}{Causal Language Modeling}
\newacronym{lms}{LMS}{Language Modeling Score}

\newcommand{\tablehead}[2]{\textbf{\shortstack[c]{\strut #1\\\strut #2}}}

\usepackage{tikz}
\usetikzlibrary{shapes.geometric, arrows, positioning, calc, arrows.meta, backgrounds, fit}
\usepackage{booktabs}  % For table rules
\usepackage{siunitx}   % For number formatting and alignment
\usepackage{tcolorbox} % For rounded corner colored boxes
\usepackage[inline,shortlabels]{enumitem}
\usepackage{subcaption}
\usepackage{pifont}% http://ctan.org/pkg/pifont
\definecolor{darkgreen}{RGB}{0,150,0}  % Custom darker green

\newcommand{\cmark}{\ensuremath{\text{\ding{51}}}}
\newcommand{\xmark}{\ensuremath{\text{\ding{55}}}}

\usepackage{mathtools}
\usepackage{amssymb}
\usepackage[normalem]{ulem}  % Paket für durchgestrichenen Text
\usepackage{marginnote}

\usepackage{colortbl}
\usepackage{tabularx}
\usepackage{bm}
\newcommand{\signG}{\mathrm{sign}_G}

\title{The \gradiend\ Python Package: An End-to-End System for Gradient-Based Feature Learning}

\author{Jonathan Drechsel  \and Steffen Herbold \\
    Faculty of Computer Science and Mathematics \\
    University of Passau \\
    \texttt{\{jonathan.drechsel,steffen.herbold\}@uni-passau.de}\\
}

\begin{document}
\maketitle
\begin{abstract}
%Interpretability experiments for language models are often implemented as ad-hoc, task-specific code, limiting reproducibility and large-scale comparison across models and features. 
We present \texttt{gradiend}, an open-source Python package that operationalizes the \gradiend\ method for learning feature directions from factual-counterfactual MLM and CLM gradients in language models. 
The package provides a unified workflow for feature-related data creation, training, evaluation, visualization, persistent model rewriting via controlled weight updates, and multi-feature comparison.
We demonstrate \texttt{gradiend} through an English pronoun running example, a semantic sentiment use case that evaluates lexical generalization to held-out target words, and a large-scale feature comparison.
%across feature classes, model families, and pruning configurations.
\end{abstract}

\glsunset{gradiend}

\section{Introduction}
\label{sec:intro}

%Interpretability studies of language models (LMs) are often implemented as task-specific research code tightly coupled to a particular dataset, model architecture, and experimental setup. 
%As a result, feature analyses are difficult to reproduce, hard to compare across models and languages, and particularly challenging to adapt or scale to new problems, feature definitions, and application settings.

Interpretability is essential for responsible deployment of modern language models, which are highly capable yet opaque \cite{molnar2025}.
Accordingly, many interpretability methods have emerged, including input attributions \cite{shap, integratedGradients}, counterfactual analyses \cite{counterfactual}, and activation-based interventions \cite{zhang2024towards}.
However, these tools largely operate at analysis time and rarely yield reusable, feature-level objects that support cross-model comparison or persistent behavioral changes.
Conversely, weight-space steering methods such as task vectors \cite{ilharco2023editing} and weight arithmetic \cite{fierro2026steering} produce persistent updates but infer directions indirectly from fine-tuning deltas (effectively accumulated gradient updates \cite{zhou2025on}).

\gradiend~\cite{gradiend} offers an interpretability-driven alternative: it learns reusable \emph{feature directions} directly from feature-specific gradient signals and instantiates them as controlled parameter-space interventions.
In this system demonstration, we present \texttt{gradiend}, an open-source Python package implementing \gradiend\  for \textbf{learning, comparing, and manipulating feature directions in language models}.
The library provides a unified, end-to-end workflow covering feature-specific data construction, feature direction learning from language model gradients, evaluation, scalable multi-feature comparison, and controlled weight updates to create \emph{persistent} rewritten model variants.
It exposes this workflow through a Hugging Face Trainer-style API, making the \gradiend\ method readily usable in practice.

\rev{
We demonstrate the package through
\begin{enumerate*}[label=(\roman*)]
    \item an English pronoun running example covering the full \texttt{gradiend} workflow,
    %\item a semantic sentiment feature with held-out target-word generalization,
    \item a semantic sentiment feature that evaluates generalization to unseen target words,
    \item large-scale feature comparisons using weight overlap and cross-encoding, and
    \item a pruning ablation quantifying efficiency gains.
\end{enumerate*}
}

\begin{figure*}[!t]
\centering
\begin{tikzpicture}[
  font=\sffamily,
  W/.store in=\W, W=2.8cm,
  H/.store in=\H, H=3.4cm,
  gap/.store in=\gap, gap=7mm,
  vgap/.store in=\vgap, vgap=3mm,
  pad/.store in=\pad, pad=2.2mm,
  % --- boxes ---
  box/.style={
    draw=black!20, rounded corners=2mm, fill=white,
    inner sep=\pad, text width=\W, align=left
  },
  topbox/.style={box, minimum height=\H},
  rightbox/.style={box},
  % --- typography ---
  boxtitle/.style={font=\bfseries\footnotesize, text=black!90},
  tiny/.style={font=\tiny, text=black!70},
  % --- API pill ---
  apipill/.style={
    draw=black!18, rounded corners=1.4mm, fill=black!3,
    inner sep=1.2mm, font=\ttfamily\tiny, text=black!70
  },
  % --- visual slot ---
  slot/.style={
    draw=black!15, rounded corners=1.6mm, fill=black!2,
    inner sep=1.8mm, align=center
  },
  % --- process arrows (block/process look) ---
  proc/.style={
    -{Triangle[length=9pt,width=12pt]},
    line width=5.2pt, draw=black!55
  }
]

% =========================
% ACCENT COLORS (one per box)
% =========================
\colorlet{cData}{blue!70!black}
\colorlet{cTrain}{violet!70!black}
\colorlet{cIntra}{green!55!black}
\colorlet{cRewrite}{orange!80!black}
\colorlet{cInter}{red!70!black}

% card border + subtle tint (ONLY color; no geometry)
\tikzset{
  boxData/.style   ={draw=cData!55,   fill=cData!6},
  boxTrain/.style  ={draw=cTrain!55,  fill=cTrain!6},
  boxIntra/.style  ={draw=cIntra!55,  fill=cIntra!6},
  boxRewrite/.style={draw=cRewrite!55,fill=cRewrite!6},
  boxInter/.style  ={draw=cInter!55,  fill=cInter!6},
  % icon coloring helper (use: icon=cData, etc.)
  icon/.style args={#1}{text=#1!85},
  % highlight helper for one element per box
  hi/.style args={#1}{fill=#1!28, draw=#1!70}
}

% global node diameter
\newcommand{\nnDiam}{1mm}

% right boxes heights so that: H = Hright + vgap + Hright
\pgfmathsetlengthmacro{\Hright}{(\H-\vgap)/2}

% =========================
% BOX LAYOUT
% =========================
\node[topbox, boxData] (B1) {};
\node[topbox, boxTrain, right=\gap of B1] (B2) {};
\node[topbox, boxIntra, right=\gap of B2] (B3) {};

% Right column: north-aligned + south-aligned to the top row
\node[rightbox, boxRewrite, minimum height=\Hright, anchor=north west]
  (B4a) at ([xshift=\gap]B3.north east) {};
\node[rightbox, boxInter, minimum height=\Hright, anchor=south west]
  (B4b) at ([xshift=\gap]B3.south east) {};

% =========================
% PROCESS ARROWS (horizontal + parallel)
% =========================
\draw[proc] (B1.east) -- (B2.west);
\draw[proc] (B2.east) -- (B3.west);

\coordinate (S4a) at (B3.east|-B4a.west);
\coordinate (S4b) at (B3.east|-B4b.west);
\draw[proc] (S4a) -- (B4a.west);
\draw[proc] (S4b) -- (B4b.west);

% =========================
% HELPERS: title + api pill + one or two visual slots
% (No \dimexpr multiplication; use "2\pad" not "2*\pad")
% =========================
\tikzset{
  boxtitle/.style={font=\bfseries\footnotesize, text=black!90},
  bodytiny/.style={font=\tiny, text=black!80, align=left},
  apipill/.style={
    draw=black!18, rounded corners=1.4mm, fill=black!3,
    inner sep=0.7mm, font=\ttfamily\tiny, text=black!70
  },
  inset/.style={
    draw=black!15, rounded corners=1.6mm, fill=black!2,
    inner sep=2.0mm, align=left
  },
  chevron/.style={-Latex, line width=0.9pt, draw=black!55}
}

% --- Helper: title + api anchored to padded corners (works best if your box uses inner xsep/ysep) ---
\newcommand{\PlaceTitleApi}[3]{%
  % Title (unchanged)
  \node[anchor=north west, boxtitle, 
    text width=2.6cm,]
    at (#1.north west) {\scriptsize #2};

  % API pill slightly outside bottom-right
  \node[
    anchor=south east,
    apipill,
  ] at ([xshift=2.5mm,yshift=-2.5mm]#1.south east) {\tiny #3};
}

\newcommand{\DataExample}[1]{%
  % Inner example box (positioned under title)
  \node[inset, anchor=north west, text width=\W-8mm] (ex#1) at ([yshift=-7mm]#1.north west) {%
    \bodytiny
    \textbf{Text:} Mary said that \texttt{[MASK]} arrived.\par\vspace{0.8mm}
    \textbf{Factual:} \texttt{[MASK]} $=$ ``she''\par
    \textbf{Counterfactual:} \texttt{[MASK]} $=$ ``he''
  };

  % Arrow / explanation line under the inner box
  \node[bodytiny, anchor=north west] (grad#1) at ([yshift=-2mm]ex#1.south west) {%
    $\rightarrow$ MLM/CLM gradients $\nabla_\theta$ \;\; $\rightarrow$ feature direction $\Delta\theta$
  };
}

% =========================
% CONTENT (minimal text, visual placeholders)
% =========================

% Data: example text (later: real mini rendered example or a tiny PDF/PNG)
\PlaceTitleApi{B1}{Feature Selection and Data Generation}{TextPredictionDataCreator()}
\node[
  anchor=north east,
  text=teal!80!black,
  scale=0.95, icon=cData,
] at ([xshift=-0.5mm,yshift=-0.0mm]B1.north east)
{\faIcon{file-alt}};
% =========================
% DATA CONTENT (modern card style)
% =========================

% Title
\node[
  anchor=north west,
  font=\tiny,
  text=black!90,
  text width=3.05cm,
  text=black!65,
] at ([yshift=-6.7mm, xshift=0mm]B1.north west)
%{Feature described by two target classes (e.g., gender: female/male)};
%{Two target classes specify \\the feature axis (e.g., \emph{female} vs.\ \emph{male} for feature \emph{gender})};
%{Two target classes specify \\the feature axis (e.g., \emph{female} vs.\ \emph{male} for feature \emph{grammatical person})};
{Two classes define the feature axis (e.g., 3rd-person singular vs.\ plural (\emph{3SG} vs.\ \emph{3PL}) for \emph{gram.\ number})};

% API pill
% Inner example card (centered with clean margins)
\iffalse
\node[
  draw=black!15,
  rounded corners=1.6mm,
  fill=black!1,
  inner sep=1mm,
  align=left,
  text=black!80,
  font=\tiny,
  text width=\W-0mm,
  anchor=north west
] (dataex) at ([yshift=-13mm, xshift=1mm]B1.north west)
{
\textbf{Text}\, 
%Mary said that \texttt{\textcolor{cData}{\textbf{[MASK]}}} arrived.\par\vspace{1mm}
The nurse Mary said that \texttt{\textcolor{cData}{\textbf{[MASK]}}} was exhausted.\par\vspace{1mm}

\textbf{Factual}\, \texttt{\textcolor{cData}{\textbf{[MASK]}}} = ``she''\par\vspace{1mm}
\textbf{Counterfactual}\, \texttt{\textcolor{cData}{\textbf{[MASK]}}} = ``he''
};
\fi
\node[
  draw=black!15,
  rounded corners=1.6mm,
  fill=black!1,
  inner sep=0.8mm,
  align=left,
  text=black!80,
  font=\fontsize{5.7pt}{6.1pt}\selectfont,,
  text width=\W-0mm,
  anchor=north west
] (dataex) at ([yshift=-15.5mm, xshift=1.3mm]B1.north west)
{
\textbf{Text}\, 
%Mary said that \texttt{\textcolor{cData}{\textbf{[MASK]}}} arrived.\par\vspace{1mm}
The nurse Mary said that \texttt{\textcolor{cData}{\textbf{[MASK]}}} was exhausted.\par\vspace{1mm}

\textbf{Factual}\, \texttt{\textcolor{cData}{\textbf{[MASK]}}} = ``she''\par\vspace{1mm}
\textbf{Counterfactual}\, \texttt{\textcolor{cData}{\textbf{[MASK]}}} = ``they''
};

% Gradient explanation line (subtle)
\node[
  anchor=north west,
  font=\tiny,
  text=black!65,
   text width=3.cm,
  align=left
] at ([yshift=-27.0mm]B1.north west)
{The MLM/CLM gradients are used as model inputs and outputs};

% GRADIEND: network diagram (later: \includegraphics)
\PlaceTitleApi{B2}{\gradiend\ Training}{Trainer.train()}
\node[
  anchor=north east,
  text=teal!80!black,
  scale=0.95,
  icon=cTrain
] at ([xshift=-0.5mm,yshift=0.1mm]B2.north east)
{\faIcon{cogs}};
%{\faIcon{project-diagram}};
%{\faIcon{chart-line}};

\begin{scope}[shift={([yshift=-1.9mm, xshift=0mm]B2.center)}, x=1mm,y=1mm]

  % x positions (simple)
  \def\xI{-8}
  \def\xF{0}
  \def\xO{8}

  % styles (match your look)
  \tikzset{
    nnode/.style={circle,draw=black!35,fill=white,minimum size=2.5mm,inner sep=0pt,line width=0.35pt},
    nhot/.style ={circle,draw=black!35,fill=cTrain!55,minimum size=2.5mm,inner sep=0pt,line width=0.35pt},
    nedge/.style={draw=black!25,line width=0.50pt},
    featArrow/.style={-Latex, draw=black!55, line width=0.45pt},
    featLabel/.style={font=\tiny, text=black!70}
  }

  % 8 y positions (top to bottom)
  \def\yone{10.5}
  \def\ytwo{7.5}
  \def\ythree{4.5}
  \def\yfour{1.5}
  \def\yfive{-1.5}
  \def\ysix{-4.5}

  % INPUT nodes I1..I8
  \node[nnode] (I1) at (\xI,\yone) {};
  \node[nnode] (I2) at (\xI,\ytwo) {};
  \node[nnode] (I3) at (\xI,\ythree) {};
  \node[nnode] (I4) at (\xI,\yfour) {};
  \node[nnode] (I5) at (\xI,\yfive) {};
  \node[nnode] (I6) at (\xI,\ysix) {};

  % FEATURE latent node
  \node[nhot] (F) at (\xF,3) {};

  % OUTPUT nodes O1..O8
  \node[nnode] (O1) at (\xO,\yone) {};
  \node[nnode] (O2) at (\xO,\ytwo) {};
  \node[nnode] (O3) at (\xO,\ythree) {};
  \node[nnode] (O4) at (\xO,\yfour) {};
  \node[nnode] (O5) at (\xO,\yfive) {};
  \node[nnode] (O6) at (\xO,\ysix) {};

  % Connections: inputs -> feature
  \foreach \i in {1,2,3,4,5,6} {
    \draw[nedge] (I\i) -- (F);
  }

  % Connections: feature -> outputs
  \foreach \j in {1,2,3,4,5,6} {
    \draw[nedge] (F) -- (O\j);
  }

  % Label "feature" with arrow to latent node
  \node[featLabel] (flab) at ([xshift=0mm,yshift=-5.2mm]F.south) {feature $h$};
  \draw[featArrow] (flab.north) -- (F);
\node[featLabel] (enc) at ([xshift=0mm,yshift=1.4mm]I1.north) {Encoder};
  \node[featLabel] (dec) at ([xshift=0mm,yshift=1.4mm]O1.north) {Decoder};

% --- Encoder/Decoder background boxes + captions ---
\begin{pgfonlayer}{background}
  % left (encoder) box
  \node[
    fit=(I1)(I6),
    inner sep=1.2mm,
    rounded corners=0.8mm,
    fill=black!3,
    draw=black!15,
    line width=0.25pt
  ] (encBox) {};

  % right (decoder) box
  \node[
    fit=(O1)(O6),
    inner sep=1.2mm,
    rounded corners=0.8mm,
    fill=black!3,
    draw=black!15,
    line width=0.25pt
  ] (decBox) {};
\end{pgfonlayer}

% captions under boxes
\node[featLabel, align=center, text=black!65, text width=1.5cm]
  at ([yshift=-3mm]encBox.south) {Encoding a scalar feature value};

\node[featLabel, align=center, text=black!65,  text width=1.5cm]
  at ([yshift=-3mm]decBox.south) { Decoding model feature update direction};

\end{scope}

% Intra Eval: encoder + decoder plots (two slots)
\PlaceTitleApi{B3}{Intra-Model Evaluation}{Trainer.evaluate()}
\node[
  anchor=north east,
  text=teal!80!black,
  scale=0.95,
  icon=cIntra
] at ([xshift=-0.1mm,yshift=-0.0mm]B3.north east)
{\faIcon{chart-bar}};

\node[
  anchor=north west,
  font=\tiny,
  text width=3.1cm,
  text=black!65,
] at ([yshift=-3mm, xshift=0mm]B3.north west)
{Encoder analysis: are feature classes separated?};

\begin{scope}[shift={([yshift=0mm, xshift=-8mm]B3.center)}]

  \colorlet{vPlus}{cIntra!75}
  \colorlet{vMinus}{cIntra!50}
  \colorlet{vZero}{cIntra!28}

  \begin{axis}[
    width=3.2cm, height=2.2cm,
    xmin=0.60, xmax=1.6,     % <-- tighter + gives left margin
    ymin=-1.05, ymax=1.05,
    axis x line*=bottom,
    axis y line*=left,
    xtick=\empty,
    ytick={-1,0,1},
    yticklabels={-1,0,+1},
    ticklabel style={font=\fontsize{4.8pt}{5.4pt}\selectfont, text=black!65},
    axis line style={draw=black!20, line width=0.3pt},
    tick style={draw=black!20, line width=0.3pt},
    clip=true,
    ylabel={$h$},
ylabel style={
  font=\fontsize{4.5pt}{5pt}\selectfont,
  text=black!70,
  yshift=-1.5mm,
},
  ]

  % split violin center line
  \addplot[draw=black!25, line width=0.25pt] coordinates {(1,-1.0) (1,1.0)};

  % LEFT HALF (+1) — peak pushed closer to +1 (bulge very near top)
  \path[draw=black!25, line width=0.25pt, fill=vPlus]
    (axis cs:1,0.50)
    .. controls (axis cs:1-0.01,0.50) and (axis cs:1-0.14,0.72) ..
    (axis cs:1-0.22,0.82)
    .. controls (axis cs:1-0.36,0.93) and (axis cs:1-0.30,0.995) ..
    (axis cs:1,1.00)
    -- (axis cs:1,0.50) -- cycle;

  % RIGHT HALF (-1) — peak pushed closer to -1 (bulge very near bottom)
  \path[draw=black!25, line width=0.25pt, fill=vMinus]
    (axis cs:1,-1.00)
    .. controls (axis cs:1+0.01,-1.00) and (axis cs:1+0.18,-0.985) ..
    (axis cs:1+0.32,-0.93)
    .. controls (axis cs:1+0.36,-0.88) and (axis cs:1+0.22,-0.68) ..
    (axis cs:1,-0.50)
    -- (axis cs:1,-1.00) -- cycle;

  % 0 violin at x=1.62 (slightly closer, to reduce empty space)
  \addplot[draw=black!25, line width=0.25pt] coordinates {(1.62,-0.5) (1.62,0.5)};

  % left half (0)
  \path[draw=black!25, line width=0.25pt, fill=vZero]
    (axis cs:1.62,-0.50)
    .. controls (axis cs:1.62-0.02,-0.50) and (axis cs:1.62-0.22,-0.25) ..
    (axis cs:1.62-0.28,0.00)
    .. controls (axis cs:1.62-0.22,0.25) and (axis cs:1.62-0.02,0.50) ..
    (axis cs:1.62,0.50)
    -- (axis cs:1.62,-0.50) -- cycle;

  \end{axis}

% --- LABELS ---

% Feature class 1 (oben links, +1 Hälfte)
\draw[black!40, line width=0.25pt]
  (rel axis cs:0.33,0.92) -- ++(2mm,0);
\node[anchor=west, font=\fontsize{4.2pt}{5pt}\selectfont, text=black!75]
  at ([xshift=1mm]rel axis cs:0.33,0.92)
  {feature class $A$};

% Feature class 2 (unten rechts, -1 Hälfte)
\draw[black!40, line width=0.25pt]
  (rel axis cs:0.6,0.08) -- ++(2mm,0);
\node[anchor=west, font=\fontsize{4.2pt}{5pt}\selectfont, text=black!75]
  at ([xshift=1mm]rel axis cs:0.6,0.08)
  {feature class $B$};

% Neutral inputs (rechtes Violin)
\draw[black!40, line width=0.25pt]
  (rel axis cs:0.96,0.55) -- ++(2mm,0);
\node[anchor=west, font=\fontsize{4.2pt}{5pt}\selectfont, text=black!75]
  at ([xshift=1mm]rel axis cs:0.96,0.55)
  {neutral};
  
\end{scope}

\node[
  anchor=north west,
  font=\tiny,
  text=black!65,
  text width=3.cm,
] at ([yshift=-18mm, xshift=0mm]B3.north west)
{Decoder analysis: change probabilities of feature tokens};

\begin{scope}[shift={([yshift=-14mm, xshift=-8mm]B3.center)}]

\colorlet{pDark}{cIntra!70}
\colorlet{pLight}{cIntra!30}

\begin{axis}[
  width=3.2cm, height=2.2cm,
  xmode=log,
  xmin=1e-6, xmax=1e-1,
  ymin=0, ymax=1,
  xlabel={learning rate},
  ylabel={$\mathbb{P}$},
  xlabel style={font=\fontsize{4.8pt}{5.3pt}\selectfont, text=black!75, yshift=3.5mm},
  ylabel style={font=\fontsize{4.8pt}{5.3pt}\selectfont, text=black!75, yshift=-1.5mm,},
  ytick={0,0.5,1},
  yticklabels={0,0.5,1},
  ticklabel style={font=\fontsize{4.0pt}{4.6pt}\selectfont, text=black!65},
  axis line style={draw=black!20, line width=0.3pt},
  tick style={draw=black!20, line width=0.3pt},
  clip=true,
  xticklabels=\empty,
]

% dark: high -> low
\addplot[
  smooth,
  mark=o,
  mark size=0.5pt,
  line width=0.7pt,
  draw=pLight
] coordinates {
  (1.000e-6, 0.95)
  (2.848e-6, 0.94)
  (8.111e-6, 0.92)
  (2.310e-5, 0.88)
  (6.579e-5, 0.80)
  (1.874e-4, 0.65)
  (5.336e-4, 0.50)
  (1.520e-3, 0.35)
  (4.328e-3, 0.20)
  (1.233e-2, 0.12)
  (3.511e-2, 0.07)
  (1.000e-1, 0.05)
};

% light: low -> high
\addplot[
  smooth,
  mark=o,
  mark size=0.5pt,
  line width=0.7pt,
  draw=pDark,
] coordinates {
  (1.000e-6, 0.05)
  (2.848e-6, 0.06)
  (8.111e-6, 0.08)
  (2.310e-5, 0.12)
  (6.579e-5, 0.20)
  (1.874e-4, 0.35)
  (5.336e-4, 0.50)
  (1.520e-3, 0.65)
  (4.328e-3, 0.80)
  (1.233e-2, 0.88)
  (3.511e-2, 0.93)
  (1.000e-1, 0.95)
};

\end{axis}
\end{scope}

\node[
  anchor=north west,
  font=\fontsize{4.5pt}{5.3pt}\selectfont,
  text=black!65,
  text width=3.cm,
] at ([yshift=-23mm, xshift=24mm]B3.north west)
{$\mathbb{P}(A)$};
\node[
  anchor=north west,
  font=\fontsize{4.5pt}{5.3pt}\selectfont,
  text=black!65,
  text width=3.cm,
] at ([yshift=-28.5mm, xshift=24mm]B3.north west)
{$\mathbb{P}(B)$};

% Deployment: rewrite visual
\PlaceTitleApi{B4a}{Model Rewrite}{Trainer.rewrite\_base\_model()}

% --- Connected Neural Network ---
% --- Visual slot container ---
\node[
  anchor=west,
  align=left,
  text width=1.4cm,   % <-- adjust width
  font=\tiny,
  text=black!65,
] (rewriteText)
at ([xshift=0mm, yshift=-1.6mm]B4a.west)
{Modify model behavior
along learned feature direction};

% --- Connected Neural Network (5 layers total) ---
\begin{scope}[shift={([yshift=-1.5mm, xshift=8mm]B4a.center)}, x=1mm,y=1mm]

  % symmetric x positions -> centered around 0
  \def\xI{-6}
  \def\xHone{-3}
  \def\xHtwo{0}
  \def\xHthree{3}
  \def\xO{6}

  \tikzset{
    nnode/.style={circle,draw=black!35,fill=white,minimum size=1.5mm,inner sep=0pt,line width=0.35pt},
    nhot/.style ={circle,draw=black!35,fill=cRewrite!55,minimum size=1.5mm,inner sep=0pt,line width=0.35pt},
    nedge/.style={draw=black!25,line width=0.30pt}
  }

  % y positions
  \def\yA{3} \def\yB{1} \def\yC{-1} \def\yD{-3}

  % INPUT (random highlights)
  \node[nnode] (I1) at (\xI,\yA) {};
  \node[nhot]  (I2) at (\xI,\yB) {};
  \node[nnode] (I3) at (\xI,\yC) {};
  \node[nhot]  (I4) at (\xI,\yD) {};

  % H1
  \node[nnode] (H11) at (\xHone,2) {};
  \node[nhot]  (H12) at (\xHone,0) {};
  \node[nnode] (H13) at (\xHone,-2) {};

  % H2
  \node[nhot]  (H21) at (\xHtwo,2) {};
  \node[nnode] (H22) at (\xHtwo,0) {};
  \node[nnode] (H23) at (\xHtwo,-2) {};

  % H3
  \node[nnode] (H31) at (\xHthree,2) {};
  \node[nnode] (H32) at (\xHthree,0) {};
  \node[nhot]  (H33) at (\xHthree,-2) {};

  % OUTPUT (random highlights)
  \node[nhot] (O1) at (\xO,\yA) {};
  \node[nnode]  (O2) at (\xO,\yB) {};
  \node[nhot] (O3) at (\xO,\yC) {};
  \node[nnode]  (O4) at (\xO,\yD) {};

  % Connections
  \foreach \i in {1,2,3,4}
    \foreach \j in {1,2,3}
      \draw[nedge] (I\i) -- (H1\j);

  \foreach \i in {1,2,3}
    \foreach \j in {1,2,3}
      \draw[nedge] (H1\i) -- (H2\j);

  \foreach \i in {1,2,3}
    \foreach \j in {1,2,3}
      \draw[nedge] (H2\i) -- (H3\j);

  \foreach \i in {1,2,3}
    \foreach \j in {1,2,3,4}
      \draw[nedge] (H3\i) -- (O\j);

\end{scope}

% --- Pen icon overlay (top-right inside slot) ---
\node[
  anchor=north east,
  text=teal!80!black,
  scale=0.95,
  icon=cRewrite
] at ([xshift=-0.5mm,yshift=-0.0mm]B4a.north east)
{\faIcon{pen}};

% Comparison: venn + heatmap (two slots)
\PlaceTitleApi{B4b}{Inter-Model Evaluation}{plot\_topk\_overlap()}
\node[
  anchor=north east,
  text=teal!80!black,
  scale=0.95,
  icon=cInter
] at ([xshift=-0.5mm,yshift=-0.0mm]B4b.north east)
{\faIcon{clone}};
%\faIcon{layer-group}
%\faIcon{clone} th

\node[
  anchor=north,
  align=left,
  text width=2.4cm,   % <-- adjust width
  font=\tiny,
  text=black!65,
] (rewriteText)
at ([xshift=-3mm, yshift=-3mm]B4b.north)
{Compare multiple features};

\begin{scope}[shift={([yshift=-3mm, xshift=-6mm]B4b.center)}, x=1mm,y=1mm]
    % --- tune ---
\def\rA{3}
  \def\rB{3}
  \def\rC{3}

  % Make A and B more similar (strong overlap), C a bit farther away
  \coordinate (A) at (-2,  0.0);
  \coordinate (B) at ( 1.2,  -0.8);
  \coordinate (C) at ( 1.5,  1.5);

  % --- base colors: 3 variants of the same family ---
  \colorlet{cA}{cInter!92!black}  % slightly darker
  \colorlet{cB}{cInter!75!black}  % medium
  \colorlet{cC}{cInter!60!black}  % lighter

  % --- fills: transparency yields overlap "combinations" automatically ---
  \def\op{0.17}

  \fill[cA, fill opacity=\op, draw=none] (A) circle (\rA);
  \fill[cB, fill opacity=\op, draw=none] (B) circle (\rB);
  \fill[cC, fill opacity=\op, draw=none] (C) circle (\rC);

  % outlines
  \draw[draw=cA!80, line width=0.6pt] (A) circle (\rA);
  \draw[draw=cB!80, line width=0.6pt] (B) circle (\rB);
  \draw[draw=cC!80, line width=0.6pt] (C) circle (\rC);

  % optional labels (tiny)
  \node[font=\tiny, text=cA!85] at ($(A)+(-3.9, 0)$) {\fontsize{4.3pt}{5.2pt}\selectfont$A \!\rightleftarrows\! B$};
  \node[font=\tiny, text=cB!85] at ($(B)+( 2, -2)$) {\fontsize{4.3pt}{5.2pt}\selectfont$B\!\leftrightarrows\! C$};
  \node[font=\tiny, text=cC!85] at ($(C)+( 2,1.8)$) {\fontsize{4.3pt}{5.2pt}\selectfont$A\!\leftrightarrows\! C$};
\end{scope}

\begin{scope}[shift={([yshift=-5.5mm, xshift=6mm]B4b.center)}, x=1mm,y=1mm,  scale=0.65]

  \def\n{6}
  \def\cell{1.7}   % unitless (mm via x=1mm)
  \def\gap{0.0}
  \def\lw{0.18}    % pt-ish; used as line width in pt below

  % derived geometry (unitless coords)
  \pgfmathsetmacro{\step}{\cell+\gap}
  \pgfmathsetmacro{\size}{\n*\step}

  % helper: i=1..6 left->right, j=1..6 top->bottom
  \newcommand{\Cell}[3]{%
    \pgfmathsetmacro{\xx}{(#1-1)*\step}
    \pgfmathsetmacro{\yy}{(\n-#2)*\step} % flip y so j=1 is TOP
    \path[draw=black!10, line width=\lw pt, fill=cInter!#3]
      (\xx,\yy) rectangle ++(\cell,\cell);
  }

  % ===== intensities (higher -> darker), symmetric pairs =====
  % A row/col
  \Cell{1}{1}{100}
 % \Cell{2}{1}{68} \Cell{1}{2}{68}
  %\Cell{3}{1}{62} \Cell{1}{3}{62}
  \Cell{4}{1}{24} \Cell{1}{4}{24}
  \Cell{5}{1}{46} \Cell{1}{5}{46}
  \Cell{6}{1}{20} \Cell{1}{6}{20}

% UL 3x3 off-diagonals (randomized but still dark) — keep symmetric
\Cell{2}{1}{44} \Cell{1}{2}{44} % A-B
\Cell{3}{1}{54} \Cell{1}{3}{54} % A-C

  % B row/col
  \Cell{2}{2}{100}
%  \Cell{3}{2}{66} \Cell{2}{3}{66}
  \Cell{3}{2}{65} \Cell{2}{3}{65} % B-C
  \Cell{4}{2}{22} \Cell{2}{4}{22}
  \Cell{5}{2}{26} \Cell{2}{5}{26}
  \Cell{6}{2}{18} \Cell{2}{6}{18}

  % C row/col
  \Cell{3}{3}{100}
  \Cell{4}{3}{21} \Cell{3}{4}{21}
  \Cell{5}{3}{23} \Cell{3}{5}{23}
  \Cell{6}{3}{17} \Cell{3}{6}{17}

  % D row/col
  \Cell{4}{4}{100}
  \Cell{5}{4}{19} \Cell{4}{5}{19}
  \Cell{6}{4}{16} \Cell{4}{6}{16}

  % E row/col (LR block darker)
  \Cell{5}{5}{100}
  \Cell{6}{5}{64} \Cell{5}{6}{64}

  % F
  \Cell{6}{6}{100}

  % outer border (robust numeric coords)
  \draw[draw=black!12, line width=0.25pt] (0,0) rectangle (\size,\size);

  % ===== labels (compute positions first; avoids "No shape named 6*(...)" errors) =====
  \pgfmathsetmacro{\yA}{(5+0.5)*\step}
  \pgfmathsetmacro{\yB}{(4+0.5)*\step}
  \pgfmathsetmacro{\yC}{(3+0.5)*\step}
  \pgfmathsetmacro{\yD}{(2+0.5)*\step}
  \pgfmathsetmacro{\yE}{(1+0.5)*\step}
  \pgfmathsetmacro{\yF}{(0+0.5)*\step}

  \node[anchor=east, text=black!60] at (1.4,\yA) {\scalebox{0.3}{$A$}};
  \node[anchor=east, text=black!60] at (1.4,\yB) {\scalebox{0.3}{$B$}};
  \node[anchor=east, text=black!60] at (1.4,\yC) {\scalebox{0.3}{$C$}};
  \node[anchor=east, text=black!60] at (1.4,\yD) {\scalebox{0.3}{$D$}};
  \node[anchor=east, text=black!60] at (1.4,\yE) {\scalebox{0.3}{$E$}};
  \node[anchor=east, text=black!60] at (1.4,\yF) {\scalebox{0.3}{$F$}};

  \pgfmathsetmacro{\xA}{(0+0.5)*\step}
  \pgfmathsetmacro{\xB}{(1+0.5)*\step}
  \pgfmathsetmacro{\xC}{(2+0.5)*\step}
  \pgfmathsetmacro{\xD}{(3+0.5)*\step}
  \pgfmathsetmacro{\xE}{(4+0.5)*\step}
  \pgfmathsetmacro{\xF}{(5+0.5)*\step}

  \node[anchor=north, text=black!60] at (\xA,1.5) {\scalebox{0.3}{$A$}};
  \node[anchor=north, text=black!60] at (\xB,1.5) {\scalebox{0.3}{$B$}};
  \node[anchor=north, text=black!60] at (\xC,1.5) {\scalebox{0.3}{$C$}};
  \node[anchor=north, text=black!60] at (\xD,1.5) {\scalebox{0.3}{$D$}};
  \node[anchor=north, text=black!60] at (\xE,1.5) {\scalebox{0.3}{$E$}};
  \node[anchor=north, text=black!60] at (\xF,1.5) {\scalebox{0.3}{$F$}};
\end{scope}
\end{tikzpicture}
\vspace{-2pt}
\caption{\gradiend\ workflow and package overview.}\label{fig:overview}
\end{figure*}

\begin{table*}[t]
\centering
\scriptsize
\setlength{\tabcolsep}{3.0pt}
\begin{tabular}{@{}lcccccc@{}}
\toprule
\textbf{System}
& \tablehead{Runtime activation}{intervention}
& \tablehead{Model-weight}{rewrite}
& \tablehead{Learns feature}{representation}
& \tablehead{Targeted feature}{learning}
& \tablehead{Feature/edit}{diagnostics}
& \tablehead{Multi-feature}{comparison} \\
\midrule
TransformerLens \cite{nanda2022transformerlens}
& \cmark & \xmark & \xmark & \xmark & \xmark & \xmark \\
\texttt{pyvene} \cite{wu-etal-2024-pyvene}
& \cmark & \xmark & \cmark & \cmark & \cmark & \xmark \\
NNsight \cite{fiotto-kaufman2025nnsight}
& \cmark & \xmark & \xmark & \xmark & \xmark & \xmark \\
EasyEdit \cite{wang-etal-2024-easyedit}
& \xmark & \cmark & \xmark & \xmark & \cmark & \xmark \\
EasyEdit2 \cite{xu-etal-2025-easyedit2}
& \cmark & \xmark & \cmark & \cmark & \cmark & \xmark \\
\texttt{steering-vectors} \cite{steeringvectors}
& \cmark & \xmark & \cmark & \cmark & \xmark & \xmark \\
SAELens \cite{saelens}
& \cmark & \xmark & \cmark & \xmark & \cmark & \xmark \\
\texttt{gradiend} (ours)
& \xmark & \cmark & \cmark & \cmark & \cmark & \cmark \\
\bottomrule
\end{tabular}
\caption{\rev{
Capability/API comparison of \texttt{gradiend} with neighboring reusable systems.
\cmark{} indicates direct support as a primary package workflow, and \xmark{} indicates no primary package support under the criterion.
}}
\label{tab:system-comparison}
\end{table*}
\iffalse
\rev{
Here, \emph{model-weight rewrite} means that the package modifies model parameters rather than only applying runtime activation interventions.
\emph{Learns feature representation} means that the package trains a reusable representation of a feature, such as a trainable intervention, steering vector, sparse feature, or \gradiend\ direction.
\emph{Targeted feature learning} means that this feature representation is learned from user-provided examples, contrasts, objectives, losses, or metrics.
\emph{Feature/edit diagnostics} means built-in evaluation or visualization of learned feature representations, interventions, edits, or rewritten models.
\emph{Multi-feature comparison} means explicit support for comparing multiple learned feature representations.
}
\fi

%\paragraph{Contributions.}
\rev{
This work introduces \texttt{gradiend} as the first reusable Python package for the \gradiend\ workflow and extends prior work \cite{gradiend, drechsel2026understandingmemorizingcasestudy} with:
\begin{itemize}
    \item new feature settings beyond the original experiments, showing that \texttt{gradiend} can be readily applied beyond the original experiments, including the learning of semantic features and cross-feature analysis;
    \item pruning utilities that reduce memory, runtime, and storage costs for large-scale analyses by improving the efficiency while still allowing effective feature learning;
    \item standardized multi-feature comparison tools, including cross-encoding, which turns comparisons between pairwise \gradiend\ models into feature-level comparisons to support reproducible and comparable research;
    \item a flexible, end-to-end Trainer-style interface for the full \gradiend\ workflow that minimizes the feature-specific implementation effort;
    \item documentation and examples that demonstrate the package on multiple feature types, including morphosyntactic and semantic features.
\end{itemize}
}

The \texttt{gradiend} is published via pypi at \url{https://pypi.org/project/gradiend}, its source code is released under the Apache 2.0 license at \url{https://github.com/aieng-lab/gradiend}, with documentation at \url{https://aieng-lab.github.io/gradiend/}.
%TODO system, demo video?

\section{Related Work}

%captum \cite{kokhlikyan2020captum}
%inseq \cite{sarti-etal-2023-inseq}
%LIT \cite{tenney-etal-2020-language}
%TransformerLens \cite{nanda2022transformerlens}
%SHAP \cite{shap}
%LIME \cite{lime}
%SAE 
%GRAIDEND \cite{gradiend}

Interpretability and model-modification systems differ in the workflows they support and the artifacts they produce. Input-attribution libraries such as SHAP \cite{shap}, Captum \cite{kokhlikyan2020captum}, \rev{and Inseq \cite{sarti-etal-2023-inseq} typically provide example or token-level explanations for specific predictions or generations.
Interactive analysis systems like LIT \cite{tenney-etal-2020-language} support dataset-level slicing and counterfactual inspection.
Mechanistic interpretability toolkits such as TransformerLens \cite{nanda2022transformerlens}, pyvene \cite{wu-etal-2024-pyvene}, and NNsight \cite{fiotto-kaufman2025nnsight} expose activations and internal components, and support activation-space interventions.
%These systems make interpretability workflows practical, but they do not provide an end-to-end workflow for learning targeted feature representations, comparing them across many features, and turning them into persistent model rewrites.
%These systems make interpretability workflows practical, but they primarily expose prediction-level explanations, activation-level analyses and interventions.
}

Sparse autoencoders (SAEs; \citealt{bricken2023monosemanticity}) extract sparse components from activation spaces and have been used to uncover interpretable features.
\rev{Reusable SAE tooling such as SAELens \cite{saelens} supports training and visualization of SAE features.}
These features are discovered unsupervised and commonly require post-hoc interpretation, and interventions are most often applied at runtime. 
Model editing methods such as ROME \cite{rome} and MEMIT \cite{meng2023massediting}, often accessed via EasyEdit \cite{wang-etal-2024-easyedit}, instead produce persistent model weight changes via targeted knowledge updates. 
\rev{Activation-steering systems such as EasyEdit2 \cite{xu-etal-2025-easyedit2} and \texttt{steering-vectors} \cite{steeringvectors} learn feature-like steering artifacts that are applied at runtime rather than as persistent model rewrites.}
Recent weight-space steering methods, such as task vectors \cite{ilharco2023editing} and weight arithmetic \cite{fierro2026steering}, can produce persistent parameter updates, but they infer directions indirectly from fine-tuning deltas \cite{zhou2025on}.

\iffalse
\gradiend\ complements these lines by supporting a feature-based workflow: learning feature directions directly from contrastive gradient signals, enabling comparison of many features at scale, and applying controlled weight updates that yield persistent rewritten models.
\rev{
Table~\ref{tab:system-comparison} summarizes this positioning of \texttt{gradiend} at the package/API level.
We include close neighboring systems for model-internal intervention, editing, steering, and feature discovery.
}\fi

\gradiend\ complements these lines by supporting a feature-based workflow: learning feature directions directly from contrastive gradient signals, enabling comparison of many features at scale, and applying controlled weight updates that yield persistent rewritten models.
\rev{
Table~\ref{tab:system-comparison} summarizes this positioning for the \texttt{gradiend} Python package, which implements \gradiend\ as an end-to-end API for feature learning, rewriting, and comparison.
We include close neighboring systems for model-internal intervention, editing, steering, and feature discovery.
}

%Representation-engineering methods such as ActAdd \cite{turner2023steering}, ITI \cite{iti}, and CAA \cite{rimsky-etal-2024-steering} identify controllable directions in activation space and intervene at inference time, but usually don't produce persistent rewritten models.

%The Learning Interpretability Tool (LIT; \citealt{tenney-etal-2020-language}) 
%LIT \cite{tenney-etal-2020-language} supports dataset-level slicing and counterfactual inspection, which is useful for testing hypotheses about phenomena such as gender, but it does not learn reusable feature objects or provide standardized tooling for parameter-space comparison of learned features.
%TransformerLens \cite{nanda2022transformerlens} enables fine-grained activation-space inspection and runtime causal interventions (e.g., activation patching), but it primarily targets analysis-time interventions rather than producing persistent, feature-aligned model variants.

\section{The \gradiend\ Python Package}
\label{sec:package}

The \gradiend\ method \cite{gradiend} learns a reusable \emph{feature direction} from factual-counterfactual gradients: an encoder compresses each gradient into a single class-separating scalar, and a decoder maps this scalar to a parameter-update direction which, when applied to the base model, yields a controlled and \emph{persistent} behavior shift along the learned feature.
\rev{
The \texttt{gradiend} package turns this workflow into a reusable interface that standardizes data creation, training, evaluation, rewriting, and feature comparison.
The package further extends the workflow, including pruning utilities and a feature-class comparison tool (cross-encoding).
}

Figure~\ref{fig:overview} summarizes the end-to-end workflow implemented by \texttt{gradiend}. 
The key design choice is to treat a \emph{feature direction} as a persistent, reusable artifact learned from factual-counterfactual model gradients: once trained, it can be evaluated, used to derive a rewritten checkpoint, and compared against other learned directions without rerunning training.
A typical run follows five steps:
\begin{enumerate}
    \item[\faIcon{file-alt}] \textbf{Feature Selection and Data Creation:} specify feature classes and construct class-specific text-prediction instances.
    \item[\faIcon{cogs}] \textbf{\gradiend\ Training:} train a \gradiend\ model on MLM or CLM gradients to obtain a feature direction.
    \item[\faIcon{chart-bar}] \textbf{Intra-Model Evaluation:} validate class separation and select rewrite hyperparameters.
    \item[\faIcon{pen}] \textbf{Model Rewrite:} export a controlled modified base model with changed feature behavior.
    \item[\faIcon{clone}] \textbf{Inter-Model Evaluation:} quantify similarity between multiple learned features.
\end{enumerate}
In the remainder of this section, we illustrate these steps with a \rev{simple English \emph{grammatical-number}} feature \rev{contrasting third-person singular pronouns (\emph{3SG}; \emph{he/she/it}) with plural pronouns (\emph{3PL}; \emph{they}).}
\rev{This running example focuses on the core workflow. Sections~\ref{sec:sentiment} and~\ref{sec:multifeature} evaluate the same interface on more complex semantic and large-scale feature settings, while additional package features are explained in the documentation.}

%\rev{This running example focuses on the core workflow. Additional features are documented in the package documentation.}

\subsection{Feature Selection and Data Generation}
\label{sec:featdef}

A feature is defined by a set of \emph{feature classes} (i.e., discrete variants of the feature, e.g.,  \emph{3SG} and \emph{3PL}). 
% A feature is defined by discrete \emph{feature classes}, such as \emph{3SG} and \emph{3PL}.
A single \gradiend\ run selects two classes and learns a direction along the induced axis between them. 
The library supports feature learning via \textbf{text prediction} objectives \rev{such as} \acrlong{mlm} (\acrshort{mlm}\glsunset{mlm}; \citealt{bert}) and \acrlong{clm} (\acrshort{clm}\glsunset{clm}; \citealt{radford2018improving}). 
Concretely, it constructs prediction instances in which a feature-specific target is to be predicted.
%For each feature class, the data module generates a class-specific text-prediction dataset with factual labels.

The \texttt{gradiend} data module centers around \texttt{TextPredictionDataCreator} and lightweight feature class specifications via \texttt{TextFilterConfig}. 
Users provide a base dataset (e.g., as a Hugging Face identifier) and one \texttt{TextFilterConfig} per feature class, defined through string-based word matches and optionally spaCy tags. % to handle syncretism (as by \citet{drechsel2026understandingmemorizingcasestudy}).

%\paragraph{Demo.}
\rev{
For the running example (\emph{3SG} vs.\ \emph{3PL}), simple word matching is sufficient because the selected pronouns are not syncretic.}
%We specify \emph{3SG} targets as \emph{he/she/it} and \emph{3PL} targets as \emph{they}. 
%Because these words are not syncretic, simple string-based word matching is sufficient. 
In addition to the class-specific datasets, \texttt{gradiend} can create feature-neutral evaluation data by excluding texts that contain any feature-specific target words.

\begin{minted}[fontsize=\scriptsize]{python}
from gradiend import TextPredictionDataCreator, TextFilterConfig
creator = TextPredictionDataCreator(
    base_data='wikimedia/wikipedia',
    hf_config='20231101.en',
    feature_targets=[
        TextFilterConfig(["he", "she", "it"], id="3SG"),
        TextFilterConfig(["they"], id="3PL"),
    ],
    min_left_context_words=10,
)
training_data = creator.generate_training_data(max_size_per_class=2500)
neutral_data = creator.generate_neutral_data(
    additional_excluded_words=["i", "we", "you"],
    max_size=1000,
)
\end{minted}

%Filtering: 130309 sent [00:36, 3590.87 sent/s, 3SG:2500 | 3PL:2500 | total=5000/5000 (100.0%)]
%2026-02-23 10:42:00 - INFO -   3SG: 2500/2500 matches (success rate: 1.92%)
%2026-02-23 10:42:00 - INFO -   3PL: 2500/2500 matches (success rate: 1.92%)
%2026-02-23 10:42:00 - INFO - Training filter stats (instances per group): {'3SG': 2500, '3PL': 2500}

\subsection{\gradiend\ Training}
\label{sec:train}

Training is exposed via a Hugging Face-\texttt{Trainer}-style interface: users specify a base model, the data, standard training parameters (e.g., batch size and learning rate) and \gradiend-specific settings (e.g., which base-model parameters are used \rev{for gradient extraction}, and which gradient variants to use as source and target signals).
%Each run produces a structured experiment directory containing checkpoints, metrics, and generated plots, which are later consumed by rewriting, and comparisons without rerunning training.
The framework supports automatic multi-seed training and best-model selection to ensure stable model convergence.
A training run produces a structured experiment directory with checkpoints, metrics, and plots for evaluation, rewriting, and comparison.

\begin{figure}[!t]
    \centering
    \includegraphics[width=\linewidth]{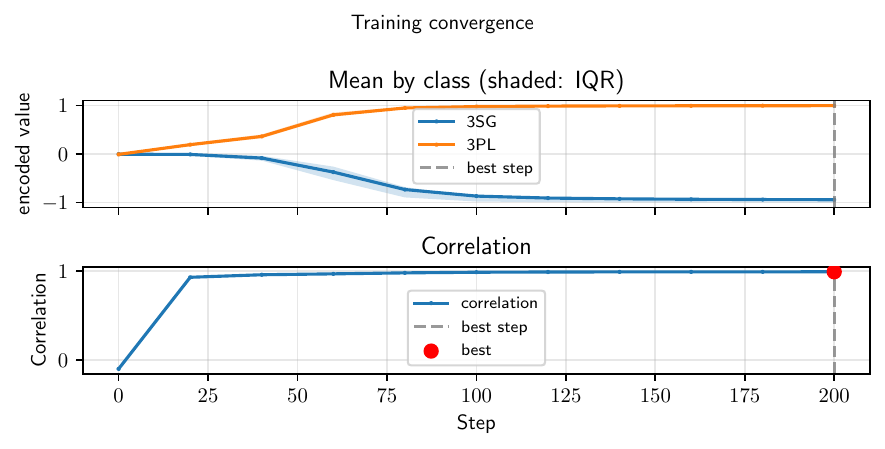}
    \vspace{-25pt}
    \caption{Training convergence example plot.}
    \label{fig:demo_training}
    % todo symbol for plots generated by library in single function call
\end{figure}

\rev{
To make large-scale studies feasible, we extend the \gradiend\ workflow with pruning.}
Pre- and post-pruning shrink the effective input and output dimensionality by removing low-importance dimensions (measured by absolute value).
Pre-pruning estimates importance from averaged gradients computed on a small random sample of training instances, while post-pruning ranks dimensions by the learned \gradiend\ parameters.
\rev{
Appendix~\ref{app:pruning} evaluates the resulting runtime, storage, and feature-quality trade-offs.
}
%To make \gradiend\ training feasible at scale, \texttt{gradiend} supports pre- and post-pruning that shrink the effective input and output dimensionality by removing low-importance dimensions. 
%Pre-pruning estimates importance from averaged gradients computed on a small random sample of training instances, while post-pruning ranks dimensions by the learned \gradiend\ parameters (using absolute magnitude as importance).
The snippet below illustrates \gradiend\ training yielding the convergence plot shown in Figure~\ref{fig:demo_training}.

\begin{minted}[fontsize=\scriptsize]{python}
from gradiend import TrainingArguments, TextPredictionTrainer, PrePruneConfig, PostPruneConfig
args = TrainingArguments(
    train_batch_size=8,
    max_steps=200,
    eval_steps=20,
    learning_rate=1e-5,
    experiment_dir="runs/demonstration",
    pre_prune_config=PrePruneConfig(n_samples=8, topk=0.1), # keep top 10% dimensions
    post_prune_config=PostPruneConfig(topk=0.01), # prune again after training; keep 1% of remaining dimensions
)
trainer = TextPredictionTrainer(
    model="bert-base-cased",
    data=training_data,
    eval_neutral_data=neutral_data,
    args=args,
)
trainer.train()
trainer.plot_training_convergence(class_spread="iqr") # Fig. 2
\end{minted}
 % todo update these plots

\subsection{Intra-Model Evaluation}
\label{sec:intramodel}

%Intra-model evaluation validates whether a \emph{single} trained \gradiend\ model yields a feature direction that is both (i) class-separating in its learned scalar feature values and (ii) usable for controlled, feature-changing rewriting of the underlying base model. 
Intra-model evaluation validates whether a \emph{single} \gradiend\ model yields a class-separating feature direction that is \rev{usable for controlled rewriting.} 
%For that, \texttt{gradiend} provides two complementary evaluation procedures: encoder evaluation, which assesses the quality of the learned feature value, and decoder evaluation, which assesses the behavioral effect of applying the decoded update.

\begin{figure}[!t]
    \centering
    \includegraphics[width=\linewidth, trim={0cm 0cm 0cm 0.5cm},
    clip]{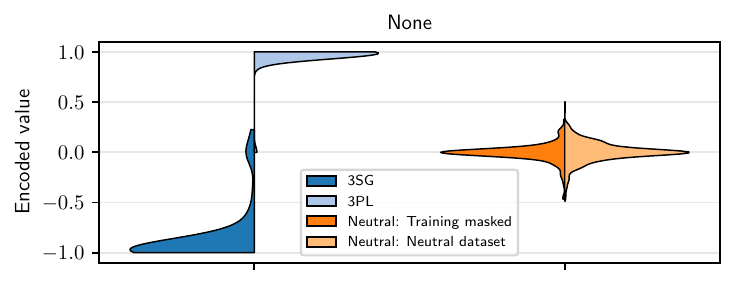}
    \caption{Encoder analysis: distribution of learned feature values across the two target classes and neutral data.}
    \label{fig:demo_encoder}
\end{figure}

%\paragraph{Encoder evaluation.}
\rev{Encoder evaluation checks whether the learned scalar feature separates the two feature classes and whether feature-neutral examples remain close to zero.
It reports separation metrics (e.g., Pearson correlation) and can generate distribution plots such as Figure~\ref{fig:demo_encoder}, showing a clear separation for the running example.
}
%Encoder evaluation analyzes which scalar values the \gradiend\ encoder assigns to gradients from different classes (e.g., the classes seen during training and neutral token-induced gradients). 
%It returns quantitative separation statistics (e.g., Pearson correlation between class labels and encoded values) and can generate distribution plots for the encoded values across class-specific and neutral data. 
%In our running example, Figure~\ref{fig:demo_encoder} shows that \emph{3SG} and \emph{3PL} instances are well separated (cluster towards $\pm 1$), while feature-neutral data concentrates near~0.

\begin{minted}[fontsize=\scriptsize]{python}
enc = trainer.evaluate_encoder(plot=True)  # Fig. 3
print("Correlation:", enc["correlation"])  # 0.9688
print("Means:", enc["mean_by_feature_class"]) # {'3PL': 0.9656, '3SG': -0.8577}
\end{minted}

\begin{figure}[!t]
    \centering
    \includegraphics[width=\linewidth]{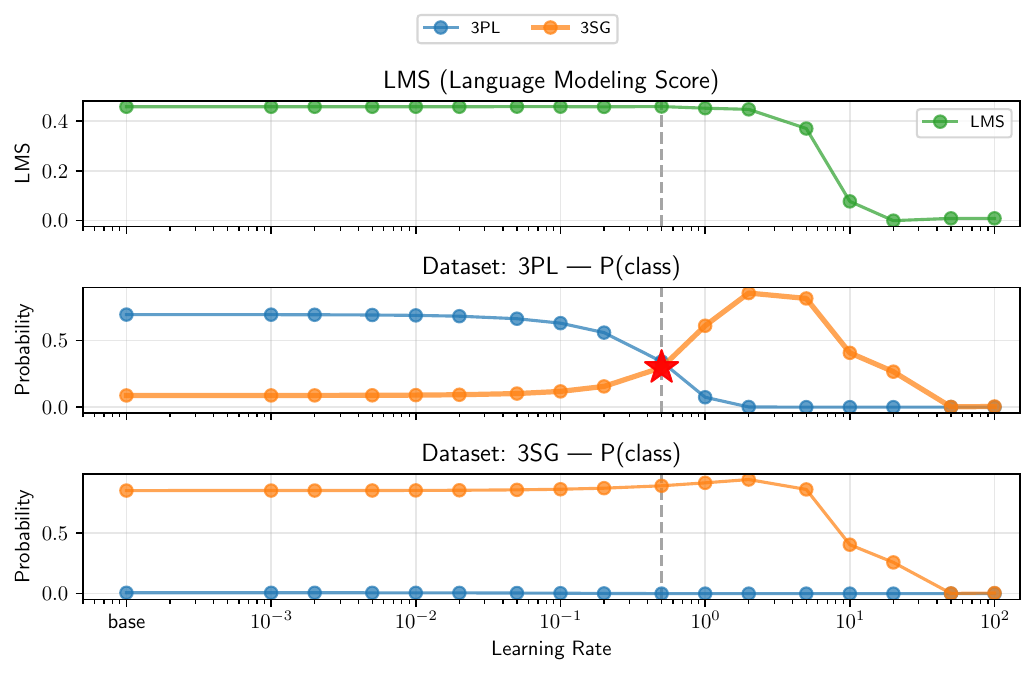}
    \caption{Decoder analysis: shift of target token probabilities; selected settings increase the target class while maintaining near-base-model \acrshort{lms}.}
    \label{fig:demo-probability-shifts}
\end{figure}

%\paragraph{Decoder evaluation.}
\rev{
Decoder evaluation assesses whether the decoded parameter update can be applied to the base model in a controlled way.}
Concretely, \texttt{gradiend} performs a small grid search over a learning rate multiplier, applies the decoded update at each scale to the base weights, and measures the induced shift in target-token probabilities.
Overly large learning rates can degrade general language modeling behavior \cite{gradiend}. Therefore, by default, \texttt{gradiend} selects the largest learning rate that satisfies a language-modeling constraint named \gls{lms}, yielding the strongest increase in the desired target-class probabilities while keeping overall model behavior close to the base model. 
Figure~\ref{fig:demo-probability-shifts} illustrates this trade-off for the \emph{3SG} target class.

\begin{minted}[fontsize=\scriptsize]{python}
dec = trainer.evaluate_decoder(plot=True, target_class="3SG")  # Fig. 4
print(dec["3SG"]["learning_rate"])  # 1.0
\end{minted}

\subsection{Model Rewrite}
\label{sec:rewrite}

\gradiend\ can export a persistently modified checkpoint whose behavior is shifted along the learned feature direction (e.g., increasing token probabilities of one target class). The rewrite can increase (default) or decrease a selected feature class, with learning rate chosen based on the previous decoder evaluation.

\begin{minted}[fontsize=\scriptsize]{python}
changed_model = trainer.rewrite_base_model(decoder_results=dec, target_class="3SG")
\end{minted}

\subsection{Inter-Model Evaluation}
\label{sec:intermodel}

\begin{figure}[!t]
    \centering
    \includegraphics[width=0.5\linewidth]{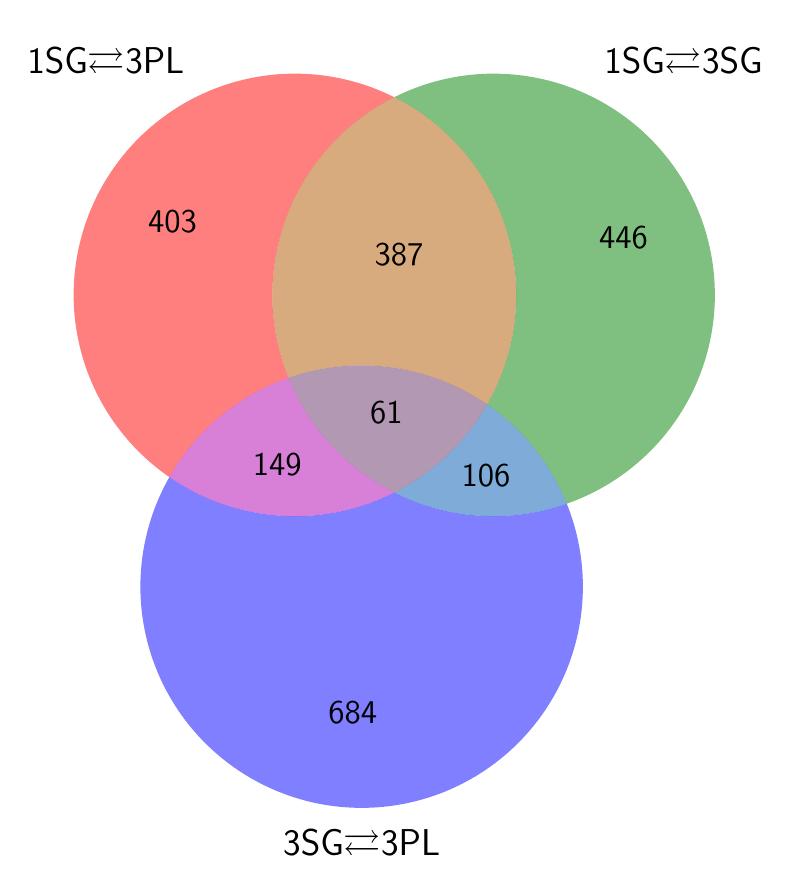}
    \vspace{-5pt}
    \caption{Venn diagram example plot.}
    \label{fig:venn}
\end{figure}

Inter-model evaluation compares \emph{multiple} trained \gradiend\ runs for the same base model to quantify how similarly different features are encoded in that model's parameters.
Because \texttt{gradiend} learns \emph{persistent} feature directions, runs can be analyzed and compared after training using only their stored artifacts, enabling scalable studies across many features.

%As a simple example, \texttt{gradiend} measures similarity via overlap of the Top-$k$ most important parameters between two \glspl{gradiend}. The high-level entrypoint \texttt{plot\_topk\_overlap} visualizes these overlaps and automatically chooses an appropriate plot type: it produces a Venn diagram for small collections of runs and falls back to a pairwise heatmap for larger collections. Section~\ref{sec:multifeature} extends this setting with cross-encoding, which aggregates comparisons over pairwise \gradiend\ models into feature-level comparisons.

\rev{
The library provides several tools for inter-model evaluation, including Top-$k$ weight overlap, cosine similarity, and cross-encoding.
As a simple example, Figure~\ref{fig:venn} visualizes the Top-$k$ overlap between three trained \gradiend\ runs as a Venn diagram.
The same comparison tools also scale to larger feature collections: Section~\ref{sec:multifeature} applies Top-$k$ overlap as a pairwise heatmap of \gradiend\ models, and uses cross-encoding to obtain a feature-class-level view.
}
% todo mention TrainerSuite?!?!?

\begin{minted}[fontsize=\scriptsize]{python}
# assuming trainers is a list of Trainer instances over the same base model
models = [trainer.get_model() for trainer in trainers]
plot_topk_overlap(models, topk=1000) # Fig. 5
\end{minted}

\section{Sentiment as a Semantic Feature}\label{sec:sentiment}

\begin{figure}[!t]
    \centering
    \includegraphics[width=0.9\linewidth]{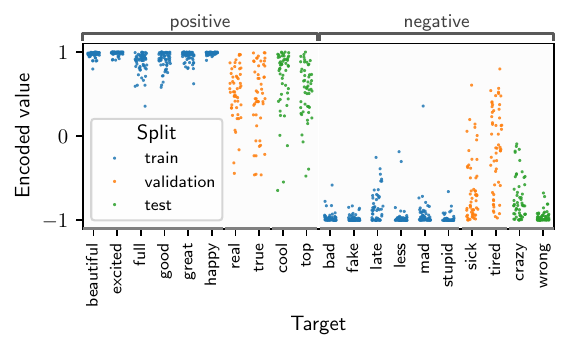}
    \vspace{-5pt}
    \caption{\rev{Sentiment encoder values grouped by target for \texttt{bert-base-cased}. %Validation and test targets are held out by vocabulary item during training.
    }}
    \label{fig:sentiment}
\end{figure}

\rev{
The running example in Section~\ref{sec:package} demonstrates \texttt{gradiend} on a \emph{morphosyntactic} pronoun feature. In prior work, we have also shown that this approach works for more complex morphosyntactic features, i.e., how German articles are determined based on noun case and gender \cite{drechsel2026understandingmemorizingcasestudy}. % todo "we" or "prior work has shown=
%To show that the same workflow also supports \emph{semantic} features, we train a sentiment \gradiend\ contrasting various positive and negative target words.
In this section, we use sentiment to demonstrate a package-level extension of the established workflow: \texttt{gradiend} can construct, split, train, evaluate, and visualize features whose classes are defined by multiple semantic target words.
We train a sentiment \gradiend\ contrasting various positive and negative target words.
Conceptually, this follows the same masked prediction setup: contexts like \emph{``The movie was [MASK]''} induce gradients for positive (e.g., \emph{great}, \emph{exciting}) and negative (e.g., \emph{bad}, \emph{sad}) targets, from which \gradiend\ learns a positive-negative feature.}

\rev{Unlike the pronoun example, sentiment provides a larger target vocabulary, i.e., various positive and negative targets. We  split the data by target word, so validation and test examples contain sentiment words not observed during training. 
This tests to what extent the learned feature captures sentiment beyond individual lexical targets, while still using the same package workflow: \texttt{gradiend} groups multiple target words into each feature class, filters and masks their occurrences, trains the positive-negative \gradiend, and produces the target-grouped visualization in Figure~\ref{fig:sentiment}. Implementation details are in Appendix~\ref{app:sentiment}. % todo check if there are implementation details
}

\rev{
Figure~\ref{fig:sentiment} shows clear separation for training targets and weaker but visible generalization to the small set of held-out target words.
The target-level view also reveals lexical differences within the sentiment classes: targets such as \emph{late}, \emph{sick}, and \emph{tired} are encoded less negatively than other negative targets in their split, potentially indicating weaker negative sentiment associations in the model.
This illustrates that \texttt{gradiend} supports semantic feature workflows beyond morphosyntactic examples, while its by-target plots can reveal lexical variation within a class.
}

\section{Large-Scale Analysis and System Validation}
\label{sec:multifeature}

\begin{figure}[!t]
    \centering
%\vspace{-10pt}    
\includegraphics[width=\linewidth]{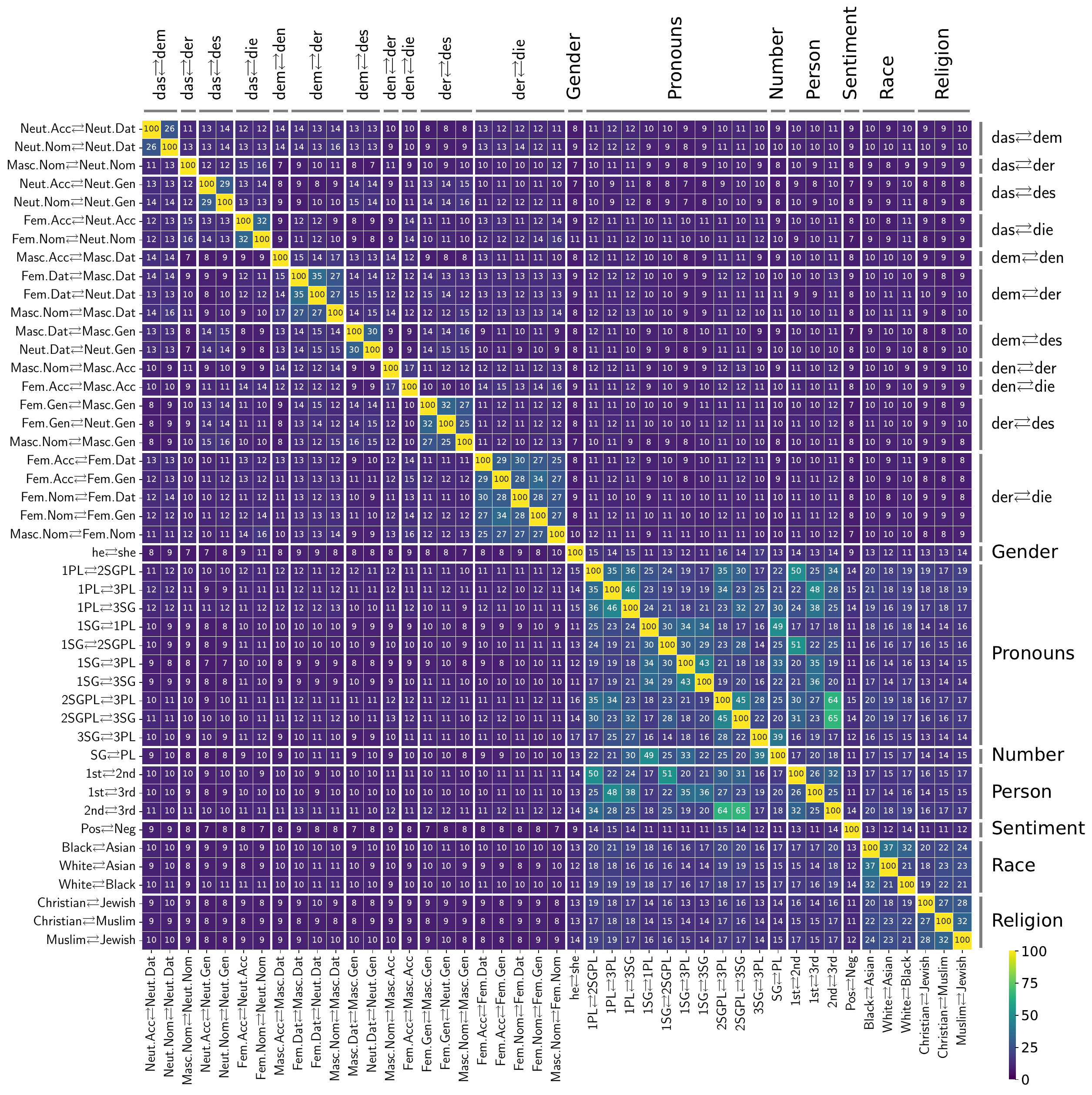}
    %\vspace{-20pt}
    \caption{Proportion of Top-$1000$ weight overlap of affected parameters. %Brighter cells indicate stronger overlap.
    } 
    \label{fig:heatmap}
\end{figure}

%Beyond single-feature evaluation, \texttt{gradiend}\ supports large-scale, multi-feature comparison by quantifying how similar different features are encoded in model parameters. 
Beyond applying \gradiend\ to individual features, this section demonstrates \texttt{gradiend}'s package-level support for large-scale, multi-feature comparison by quantifying how similarly different features are encoded in parameters of the base model.
\rev{
We evaluate this functionality on a heterogeneous set of features using the multilingual model \texttt{google-bert/bert-base-multilingual-cased} \cite{bert}.}
%Figure~\ref{fig:heatmap} summarizes this analysis across heterogeneous features. 
The analysis includes reproduced use-cases from prior work, namely English gender (\emph{he}/\emph{she}), race (\emph{Asian/Black/White}), religion (\emph{Christian/Jewish/Muslim}) \cite{gradiend}, and German gender-case articles \cite{drechsel2026understandingmemorizingcasestudy}. 
\rev{We further include the English pronoun paradigm from Section~\ref{sec:package}, extended to all five pronoun classes:
\emph{1SG} (\emph{I}), \emph{1PL} (\emph{we}), \emph{2SGPL} (\emph{you}), \emph{3SG} (\emph{he/she/it}), and \emph{3PL} (\emph{they}).}
From these classes, the framework can not only train each pair of these, but also combine features together to derive more general grammatical number and person features. Using the \texttt{class\_merge\_map} of the Trainer, feature classes can be merged together such as the singular instances \emph{1SG} and \emph{3SG} as well as plural classes \emph{1PL} and \emph{3PL} to derive a general \emph{grammatical number} feature.
\rev{Finally, we include the sentiment feature from Section~\ref{sec:sentiment} (\emph{Pos}/\emph{Neg}), yielding a collection of morphosyntactic, semantic, and social features.
}

\begin{figure}[!t]
    \centering
%\vspace{-10pt}    
\includegraphics[width=\linewidth]{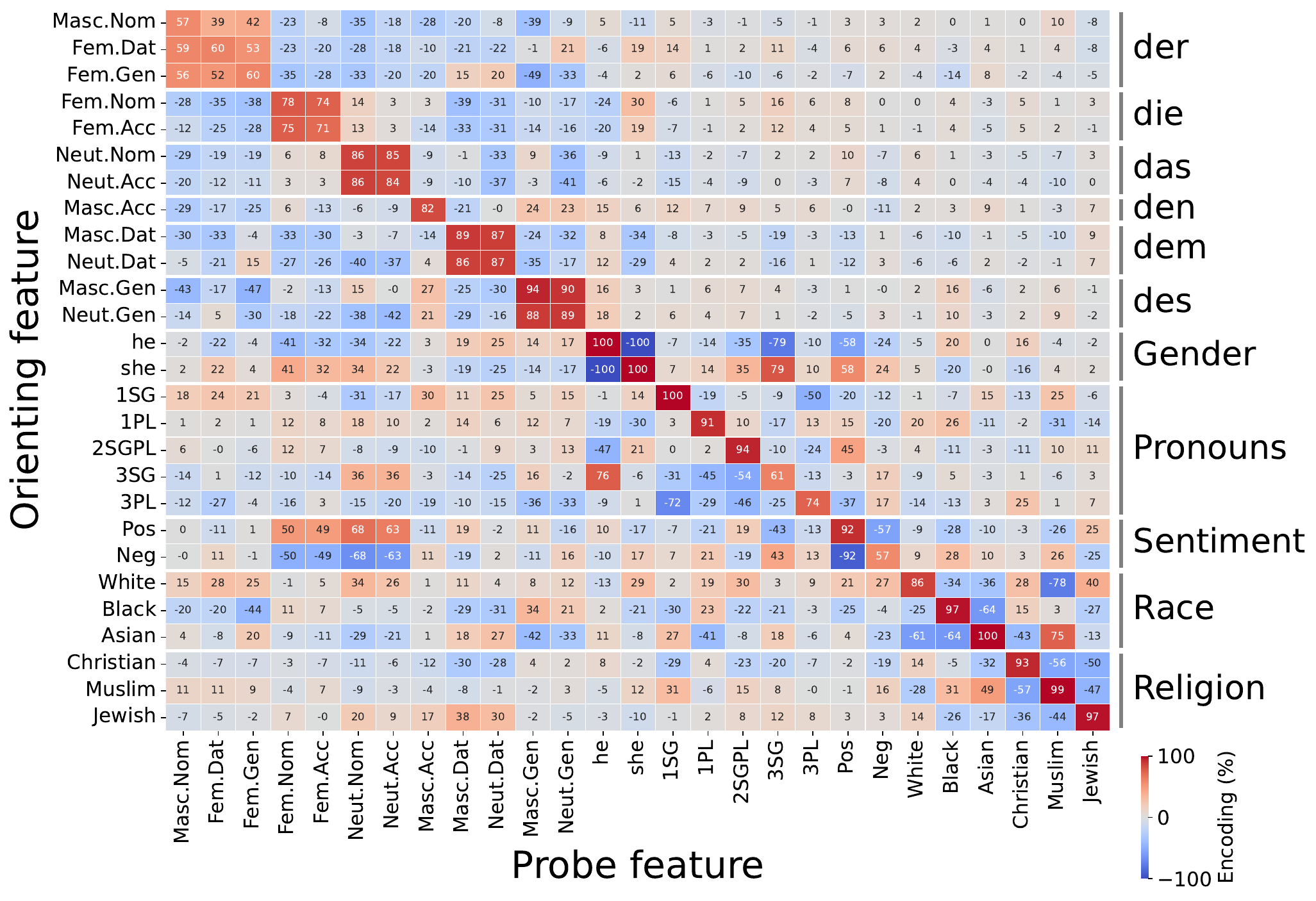}
  %  \vspace{-20pt}
   \caption{\rev{Cross-encoding matrix. Rows are orienting features, i.e., the feature classes used to align pairwise \gradiend\ models. Columns are probe features whose examples are encoded.}}
    \label{fig:cross-encoding}
\end{figure}

\rev{
The analysis uses \texttt{gradiend}'s multi-seed mode, which aggregates results over convergent random seeds to obtain more stable estimates.
The subsequent large-scale analyses report means over three convergent seeds.}

\rev{Figure~\ref{fig:heatmap} reports proportional overlap of the Top-$k$ most affected parameters ($k{=}1000$) between 45 pairwise \gradiend\ runs trained over pairs of feature classes described above.
%This pairwise overlap visualization extends the small-set Venn diagrams used in prior case studies \cite{drechsel2026understandingmemorizingcasestudy} into a large-scale feature analysis.
This package-level visualization scales the small-set Venn diagrams used in prior case studies \cite{drechsel2026understandingmemorizingcasestudy} to large feature collections.}
The block structure around the diagonal indicates that related features often share a non-trivial intersection of affected parameters.
In particular, German article transitions that share the same surface form exhibit higher overlap, consistent with the syncretism-driven analysis by \citet{drechsel2026understandingmemorizingcasestudy}.
Cross-group overlap is generally smaller, although race and religion show comparatively larger overlap than most other cross-family pairs.

\rev{As an additional package-level comparison tool, we complement this weight-space view by introducing an oriented cross-encoding matrix, formalized in Appendix~\ref{app:cross-encoding}.
For each feature class, cross-encoding takes the \gradiend\ models that contain this class, aligns their signs toward it, and measures how strongly they encode probes from the other feature classes.
The two analyses are complementary: Top-$k$ overlap compares pairwise \gradiend\ features by their affected parameters, whereas cross-encoding asks how \gradiend\ features encode probes from another. 
Thus, the former is a direct and inexpensive comparison of trained models, while the latter requires additional encoder evaluations but yields a feature-class-level view that is easier to interpret.
} 

\rev{The feature-level matrix in Figure~\ref{fig:cross-encoding} reveals cross-domain associations that are less direct to read from the pairwise overlap matrix. 
For example, \emph{Muslim} is positively associated with \emph{Asian} and \emph{Black}, negatively associated with \emph{White}, and is the only religion feature with a clearly negative (\emph{Neg}) sentiment association. 
In contrast, \emph{Christian} and \emph{Jewish} are more positively aligned with \emph{White}, while only the race feature \emph{Asian} is positively aligned with \emph{Muslim}. 
These patterns suggest that the model may encode \emph{Muslim} in a racialized and comparatively negative way. %, also based on the \emph{Pos/Neg} encodings.
Separately, English pronoun gender (\emph{he}/\emph{she}) is associated with German grammatical gender in some cases (\emph{Masc.}/\emph{Fem.}), indicating that social and grammatical gender are not fully separated in the representation.
%We interpret these patterns as examples of hypotheses surfaced by the package rather than as conclusive findings. 
}
\rev{
To scope these interpretations, we report seed-level standard deviations in Figure~\ref{fig:cross-encoding-std}. Since no null controls are included, we interpret the patterns as hypotheses surfaced by the package rather than as conclusive findings.
}

Overall, the analysis illustrates a central strength of \texttt{gradiend} as a package: feature directions can be trained, compared, and analyzed at scale through a unified interface, enabling both replication of prior findings and rapid exploration of new hypotheses.

\section{Conclusions and Future Work}
\label{sec:conclusion}

We presented \texttt{gradiend}, an open-source Python package that makes the \gradiend\ method available as a reusable, end-to-end workflow for learning, evaluating, rewriting, and comparing feature directions in language models. By treating learned feature directions as persistent artifacts and providing standardized utilities for intra-model evaluation and inter-model comparison, the toolkit supports reproducible analyses that scale from a single feature experiment to large, multi-feature studies.
\rev{
In this work, we further introduced pruning for efficient large-scale training and feature-comparison tools such as cross-encoding, which moves beyond pairwise \gradiend\ comparisons toward feature-level analysis, and demonstrated their utility on new use cases.
}

%At present, \texttt{gradiend} supports feature learning from text-prediction gradients based on \gls{mlm} and \gls{clm} gradients. 
%A natural next step is to extend the framework to additional gradient-producing objectives and modalities, such as text classification (e.g., sequence or token classification gradients) and non-text domains including vision models. 

\bibliography{custom}

\appendix

\newpage
%\clearpage

\section{Package Implementation Details}
\label{sec:engineering}

The codebase is organized into five main modules: \texttt{data}, \texttt{model}, \texttt{trainer}, \texttt{evaluator}, and \texttt{visualizer}. 
Most workflows are accessible through the \texttt{trainer} abstraction, which exposes training, evaluation, and visualization while keeping low-level model internals optional. For custom analyses, \gradiend\ model objects can be accessed via \texttt{Trainer.get\_model()}.

The implementation separates task-specific data and training logic from the core components via task-specific abstractions such as \texttt{TextPredictionTrainer}. % (as a subclass of \texttt{Trainer}).
This design makes it straightforward to add new gradient-producing objectives and modalities with minimal code extensions, e.g., vision models.

The default package installation via \texttt{pip install gradiend} builds on core libraries such as \texttt{torch} and \texttt{transformers} while keeping additional dependencies minimal to reduce version conflicts. This core setup supports data generation, training, evaluation, and model rewriting.
Additional functionality can be enabled using: % through optional dependency groups:
\begin{itemize}
    \item \texttt{gradiend[data]}: installs libraries for advanced dataset access and generation, including \texttt{datasets} and \texttt{spacy}.
    \item \texttt{gradiend[plot]}: installs visualization libraries such as \texttt{matplotlib} and Venn diagram utilities.
    \item \texttt{gradiend[recommended]}: installs both \texttt{[data]} and \texttt{[plot]}, and adds \texttt{safetensors}, which is automatically used for safer checkpoint serialization when available. % todo accelerate?
\end{itemize}

\section{Sentiment Implementation Details}
\label{app:sentiment}

\rev{
For the sentiment experiment in Section~\ref{sec:sentiment}, we construct masked sentences from \texttt{tweet\_eval} sentiment tweets \cite{rosenthal2017semeval, barbieri2020tweeteval} and the NRC Emotion Lexicon \cite{Mohammad13}.
We use the positive and negative lexicon categories, restrict candidate targets to adjectives occurring in the tweet corpus, canonicalize them by lowercasing and lemmatization, and retain the ten most frequent words per class, defining the feature classes \emph{Pos} and \emph{Neg}.
For each selected target occurrence, \texttt{gradiend} replaces the target word with the mask token and uses the original word as the prediction label.
To evaluate lexical generalization, we split the data by canonical target word (by setting \texttt{split\_col="heldout"}) with a 60/20/20 train/validation/test split. Hence, all examples of a given target word occur in exactly one split, so validation and test targets are unseen during training.
The target-grouped visualization in Figure~\ref{fig:sentiment} is produced from the encoder evaluation output using \texttt{trainer.plot\_encoder\_by\_target()}.
}

\iffalse
\begin{minted}{python}
feature_targets = [
    TextFilterConfig(targets=positive_words, id="positive", spacy_tags={"pos": "ADJ"}, use_lemma=True),
    TextFilterConfig(targets=negative_words, id="negative", spacy_tags={"pos": "ADJ"}, use_lemma=True),
]
\end{minted}
\fi

\begin{figure}[!t]
    \centering
    \includegraphics[width=0.9\linewidth]{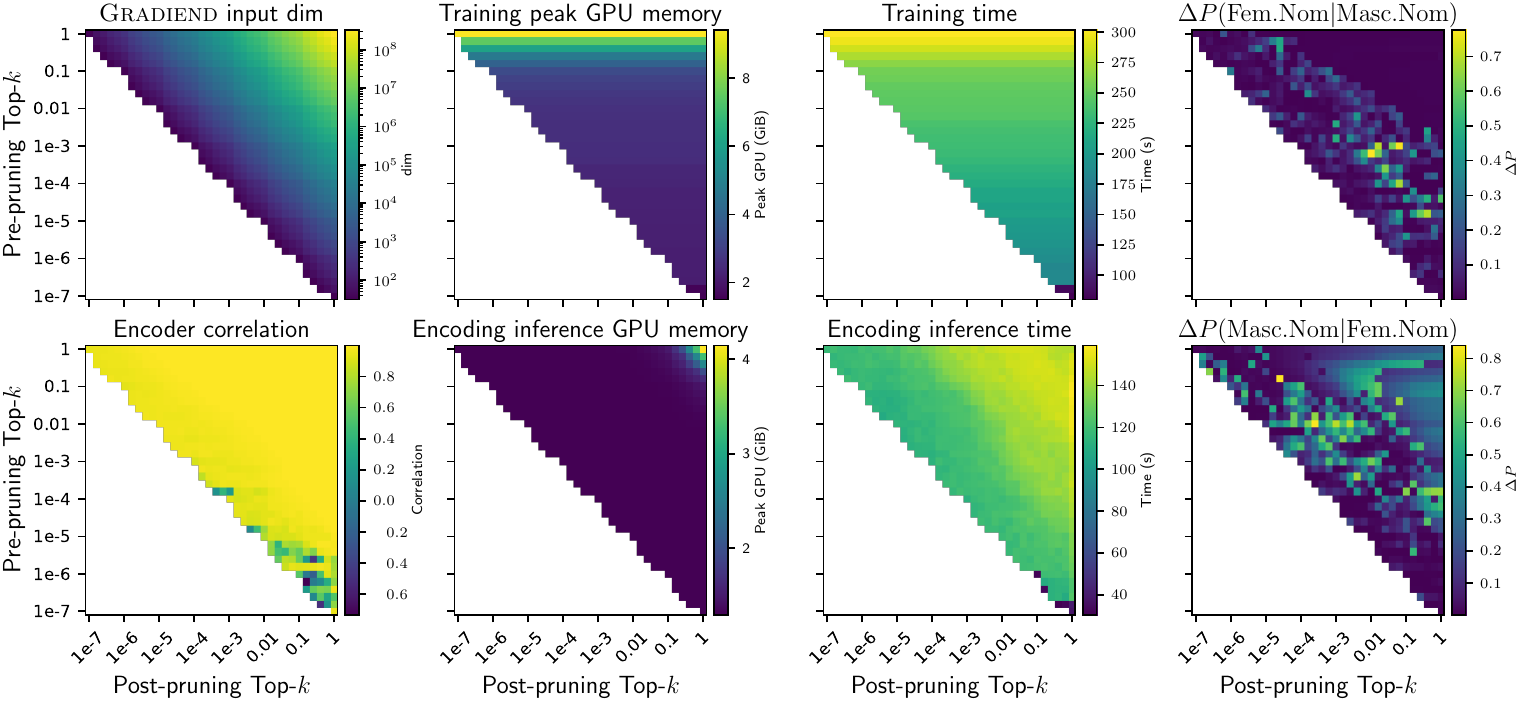}
    \caption{\rev{Pruning ablation including cost, encoder quality, and decoder shifts across pre-/post-pruning values. $\Delta P(x|y)$ denotes probability of target $x$ on data $y$.}}
    \label{fig:pruning}
    \vspace{-5pt}
\end{figure}

\section{Pruning Ablation}\label{app:pruning}

\rev{
%We evaluate pruning using the \emph{Fem.Nom.} vs.\ \emph{Masc.Nom.} feature as an example, measuring its effect on resource usage, encoder quality, decoder-induced probability shifts, and weight-overlap stability.
Using the \emph{Fem.Nom.} vs.\ \emph{Masc.Nom.} feature as an example, we evaluate how pruning affects resource usage, encoder quality, decoder-induced probability shifts, and weight-overlap stability.
Figure~\ref{fig:pruning} shows that pre-pruning, which reduces the effective \gradiend\ dimensionality before training, yields most training-memory savings already at Top-$k{=}0.1$, since gradient computation still requires the full base model (unless late layers are fully pruned).
Post-pruning only affects post-training analysis such as encoding.
Memory savings can be substantial (about a factor of three) while training time improves less strongly.
Encoder correlations and decoder shifts remain stable except under very aggressive pruning. Decoder shifts vary non-smoothly due to grid-based learning rate selection, but non-aggressive pruned settings often reach at least the unpruned probability shift.
}

\begin{figure}[!t]
    \centering
    \includegraphics[width=0.9\linewidth]{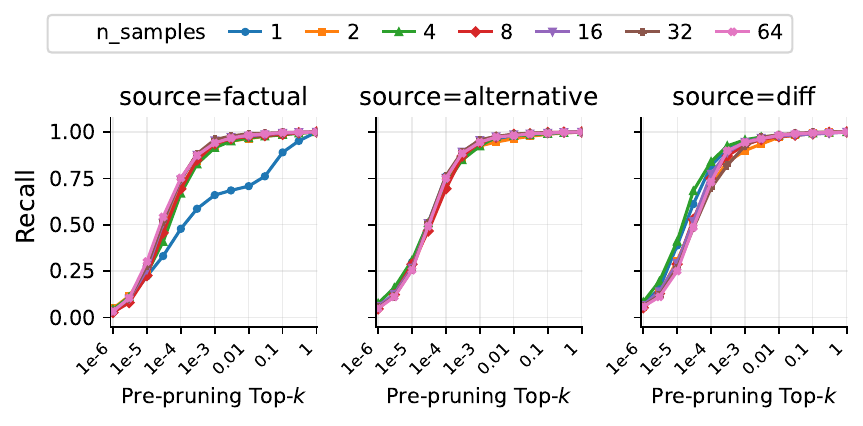}
    \vspace{-5pt}
    \caption{\rev{Pre-pruning ablation: recall of the unpruned Top-$1000$ affected parameters.}}
    \label{fig:prepruning}
\end{figure}

\rev{
Figure~\ref{fig:prepruning} complements this analysis by asking whether the resource savings from pre-pruning come at the cost of less stable weight-based comparisons.
It measures how well different pre-pruning settings recover the unpruned Top-$1000$ affected parameters.
Recall remains high for moderate pre-pruning: Top-$k{=}0.1$ is close to the unpruned reference across all gradient sources and sample counts, and Top-$k{=}0.01$ retains at least $90\%$ recall in most settings.
Lower sample counts can reduce recall, while using more than eight samples yields little additional benefit for Top-$k\geq 0.01$.
The gradient source usually has only a small effect on recall.
Although this ablation is limited to one feature and seed, it suggests that \texttt{source=alternative}, \texttt{topk=0.1}, and \texttt{n\_samples=8} are reasonable provisional package defaults: \texttt{source=alternative} performs similarly to \texttt{source=diff} here, but avoids computing both factual and alternative gradients.
}

\rev{
More generally, this ablation suggests that moderate pre-pruning may be appropriate when weight-overlap analyses should remain close to the unpruned reference (Figure~\ref{fig:prepruning}), whereas more aggressive pruning may still be acceptable when the analysis only relies on encoder evaluation or rewriting (Figure~\ref{fig:pruning}).
}

\section{Cross-Encoding Matrix}
\label{app:cross-encoding}

\rev{
Let $\mathcal{F}\coloneqq\{f_1,\ldots,f_K\}$ be feature classes  (e.g.\ \emph{3SG}, \emph{Fem.Nom.}) and $\mathcal{G}\coloneqq\{G_1,\ldots,G_M\}$
a set of pairwise \gradiend\ models (e.g., \emph{3SG}$\rightleftarrows$\emph{3PL}), each trained on an ordered pair $(a_G,b_G)\in\mathcal{F}^2$.
We evaluate each $G$ on a shared test pool $\mathcal{D}$.
%Each probe $x\in\mathcal{D}$ has a factual feature class $\mathrm{fac}(x)\in\mathcal{F}$.
%Each probe $x\in\mathcal{D}$ has a factual feature class $\mathrm{fac}(x)\in\mathcal{F}$ and a counterfactual feature class $\mathrm{cnf}(x)\in \mathcal{F}$.
%For a probe feature $f_j$, let $\mu_G(f_j)$ be the mean encoder value of $G$ over all probes with $\mathrm{fac}(x)=f_j$.
Each probe $x\in\mathcal{D}$ is associated with a factual class $\mathrm{fac}(x)\in\mathcal{F}$ and a counterfactual class $\mathrm{cnf}(x)\in\mathcal{F}$.
%For a probe feature $f_j$, let $\mu_G(f_j)$ be the mean encoder value of $G$ over all probes, such that if the \gradiend\ model uses factual gradients as input, this is over all $fac(x)=f_j$, and if the \gradiend\ model uses counterfactual gradients as input, this is over $\cnf(x)=f_j$.
For a probe feature $f_j$, let $\mu_G(f_j)$ be the mean encoded value of $G$ over all probes whose \gradiend\ source-side class is $f_j$, i.e., $\mathrm{fac}(x)=f_j$ if $G$ uses factual gradients as input, and $\mathrm{cnf}(x)=f_j$ if $G$ uses counterfactual gradients as input.
}

\rev{
For each entry $M_{i,j}$ of the cross-encoding matrix $M$, $f_i$ is the \emph{orienting feature} and $f_j$ is the \emph{probe feature}: we ask whether \gradiend\ models that separate $f_i$ from other features assign high encoder values to $f_j$ probes.
%The row feature $f_i$ therefore determines which pairwise models are averaged and how their signs are aligned, and the column feature $f_j$ determines which probes are encoded.
Let $\mathcal{G}_{f_i}=\{G:f_i\in\{a_G,b_G\}\}$ and define
$\signG(f_i)=-1$ if $f_i=a_G$ and $+1$ if $f_i=b_G$.
The cross-encoding matrix is then defined as
\[M_{i,j} \coloneqq \frac{1}{|\mathcal{G}_{f_i}|} \sum_{G\in\mathcal{G}_{f_i}} \signG(f_i)\,\mu_G(f_j).\]
The matrix is generally not symmetric, and hence not a classical similarity matrix: rows aggregate models oriented toward a feature, whereas columns select the probes being encoded.
Diagonal entries reflect self-encoding rather than perfect self-similarity and are therefore typically below $1$.
}

\section{\rev{Supplementary Cross-Encoding Analyses}}
\label{app:decoder-only}

\begin{figure}[!t]
    \centering
    \includegraphics[width=\linewidth]{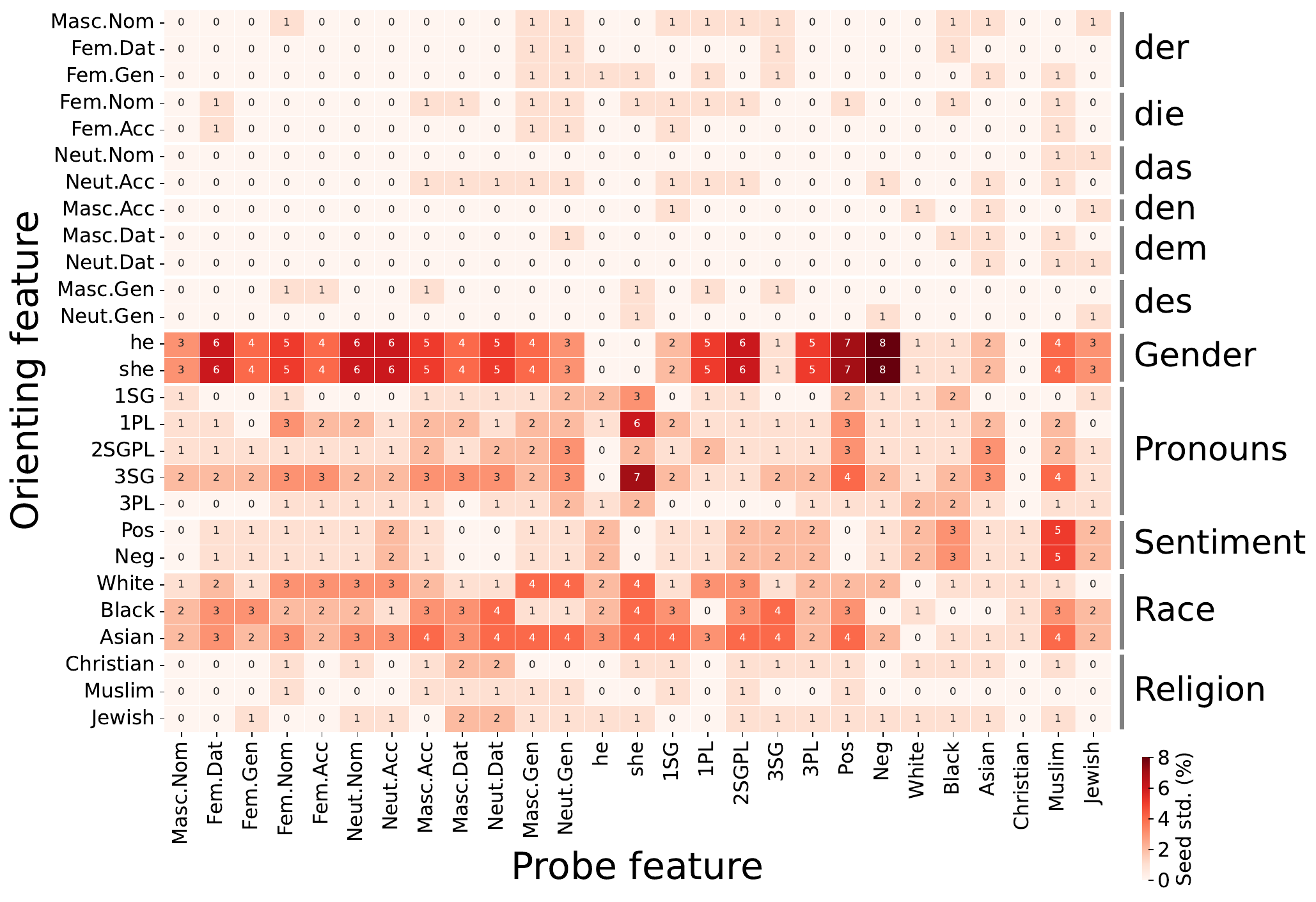}
    \vspace{-12pt}
    \caption{\rev{Seed-level standard deviation of the cross-encoding matrix in Figure~\ref{fig:cross-encoding}.}}
    \vspace{5pt}
    \label{fig:cross-encoding-std}
\end{figure}

\rev{
Figure~\ref{fig:cross-encoding-std} reports seed-level standard deviations for the cross-encoding matrix in Figure~\ref{fig:cross-encoding}.
The values are generally small relative to the main cross-encoding scale, suggesting that the visualization is not dominated by seed variation.
However, this stability check does not replace null controls, so the cross-domain associations in Section~\ref{sec:multifeature} remain exploratory.
}

\begin{figure}[!t]
    \centering
    \includegraphics[width=\linewidth]{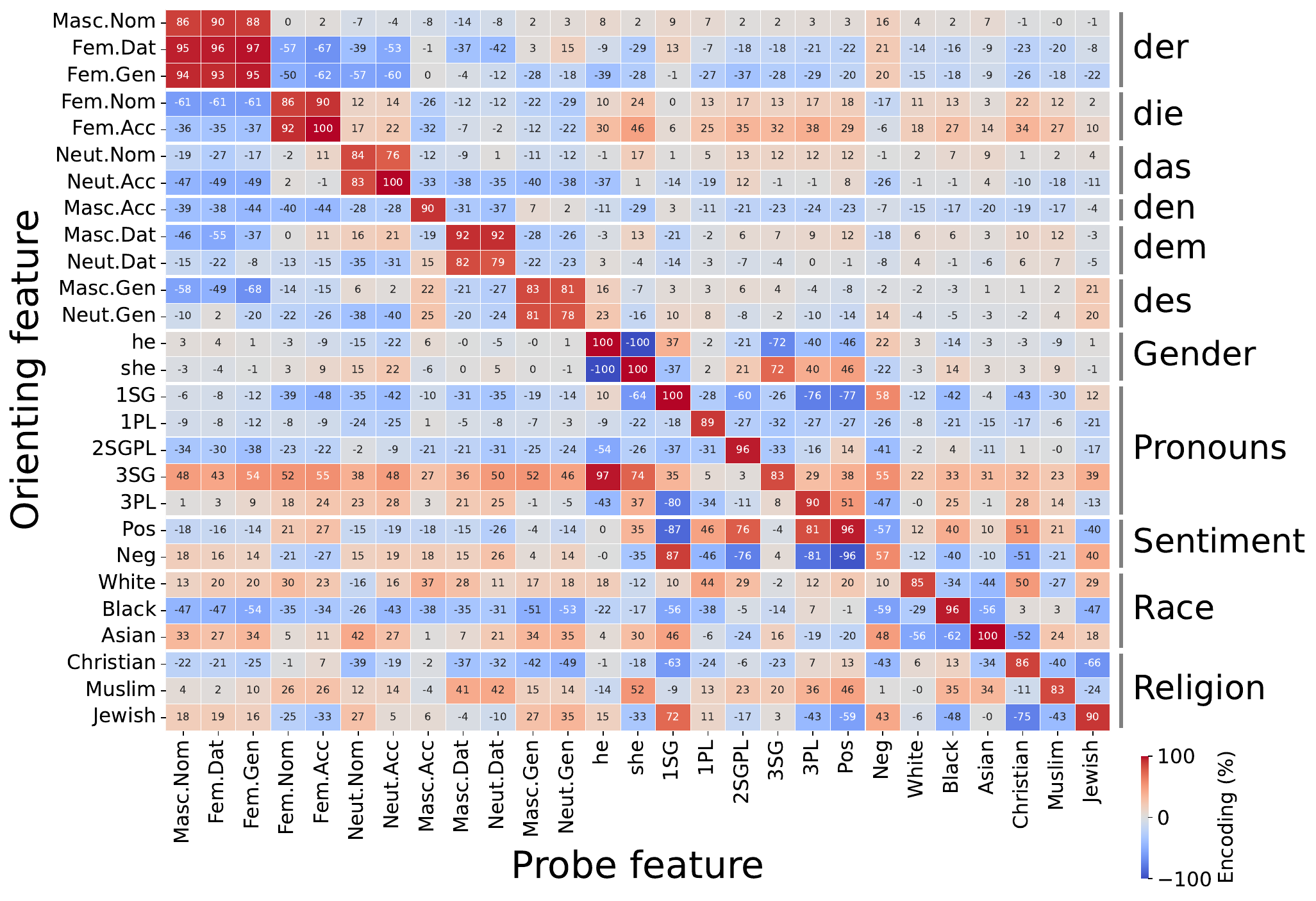}
    \vspace{-10pt}
    \caption{\rev{Cross-encoding matrix fo Qwen-2.5.}}
    \label{fig:cross-encoding-decoder}
\end{figure}

\rev{
To evaluate whether the large-scale comparison workflow also applies beyond encoder-only masked language models, we repeat the cross-encoding analysis from Section~\ref{sec:multifeature} with Qwen-2.5 0.5B \cite{qwen2, qwen2.5} using Causal Language Modeling (CLM) gradients \cite{radford2018improving}.
Figure~\ref{fig:cross-encoding-decoder} shows the resulting heatmap.
}

\rev{
%Most feature directions converge under this setting, demonstrating that the package workflow also supports decoder-only models.
%However, the full sentiment feature from Section~\ref{sec:sentiment} did not converge, indicated by the symbol \textdagger, which \texttt{gradiend} adds automatically in such plots.
%For this decoder-only replication, we therefore also include a simpler \emph{good} vs.\ \emph{bad} sentiment contrast, which converges.
}

\end{document}